\documentclass[10pt,twocolumn,letterpaper]{article}

\usepackage[accsupp]{axessibility} 
\usepackage{bigfoot}
\usepackage{multirow}
\usepackage{blindtext}

\usepackage{wacv}              

\definecolor{wacvblue}{rgb}{0.21,0.49,0.74}
\usepackage[pagebackref,breaklinks,colorlinks,allcolors=wacvblue]{hyperref}

\def\wacvPaperID{210} 
\def\confName{WACV}
\def\confYear{2026}

\newcommand{\refdiffuser}{{\scshape RefDiffuser }}

\title{Conditional Text-to-Image Generation with Reference Guidance}

\author{
Taewook Kim$^{1}$,
Ze Wang$^{2}$,
Zhengyuan Yang$^{3}$,
Jiang Wang$^{3}$,
Lijuan Wang$^{3}$,
Zicheng Liu$^{2}$,
Qiang Qiu$^{1}$ \\[1em]
$^{1}$Purdue University \quad
$^{2}$AMD \quad
$^{3}$Microsoft \\[0.5em]
}

\begin{document}
\twocolumn[{
\renewcommand\twocolumn[1][]{#1}
\maketitle
\includegraphics[width=1\textwidth]
{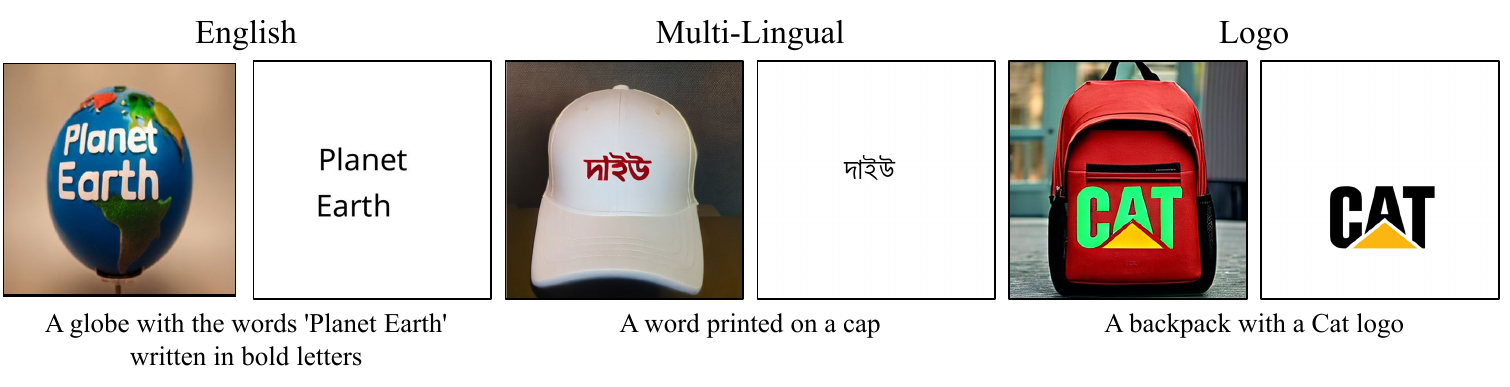}
\captionof{figure} {\textbf{Samples produced by our method.} The proposed \refdiffuser generates accurate English scene-text images, Multi-lingual scene-text images, and Logo images based on reference conditions.}
\vspace{2em}
\label{fig:teaser}
}]
\def\thefootnote{}\footnotetext{Contact: \texttt{kim3803@purdue.edu}}\def\thefootnote{\arabic{footnote}}

\begin{abstract}
Text-to-image diffusion models have demonstrated tremendous success in synthesizing visually stunning images given textual instructions. Despite remarkable progress in creating high-fidelity visuals, text-to-image models can still struggle with precisely rendering subjects, such as text spelling. 
To address this challenge, this paper explores using additional conditions of an image that provides visual guidance of the particular subjects for diffusion models to generate.
In addition, this reference condition empowers the model to be conditioned in ways that the vocabularies of the text tokenizer cannot adequately represent, and further extends the model's generalization to novel capabilities such as generating non-English text spellings. We develop several small-scale expert plugins that efficiently endow a Stable Diffusion model with the capability to take different references. Each plugin is trained with auxiliary networks and loss functions customized for applications such as English scene-text generation, multi-lingual scene-text generation, and logo-image generation. Our expert plugins demonstrate superior results than the existing methods on all tasks, each containing only 28.55M trainable parameters.
\end{abstract}

\section{Introduction}
\label{sec:intro}
Recent developments in text-to-image models have yielded groundbreaking achievements, showcasing an unprecedented capability in translating natural language descriptions into visually compelling depictions. Despite the remarkable capabilities of generating images given general textual instructions, off-the-shelf text-to-image models usually struggle with precisely generating specific subjects at a high standard. For example, Stable Diffusion (SD) \cite{SD} is known for English misspelling. 

Several research works have sought to address this issue by substituting conventional CLIP text encoders with a more powerful language model, such as T5 \cite{t5}. As demonstrated by DeepFloyd \cite{deepfloyd}, Imagen \cite{imagen} and eDiff-I \cite{ediffi}, large language models enhance both sample fidelity and image-text alignment. While scaling up language models can enhance language comprehension, it comes at a substantial cost of computation and model training. Another line of research strives to resolve this challenge with additional conditions that exhibit spatial correlation with the input. TextDiffuser~\cite{textdiffuser} utilized a segmentation mask predicted by a layout generation module as an additional condition. In a similar vein, GlyphDraw~\cite{glyphdraw} proposed using text renderings of target keywords to condition the model for generating Chinese scene-text images. However, these methods demonstrate limited capability in generating images that are beyond the vocabulary of the text encoder they are trained with. For instance, modern text-to-image models universally employ English language models \cite{clip} for prompt embedding, hindering their generalization to image generation with text from a different language (\Cref{fig:teaser}, middle) or accurate rendition of non-text subjects such as logos (\Cref{fig:teaser}, right).


To resolve the aforementioned shortcoming of text-to-image diffusion models,
we propose \refdiffuser, an approach that uses reference images as an additional source of condition, improving the generation results of particular subjects of interest with the guidance of explicit visual reference of their appearance. 
Our method is built upon the SD \cite{SD}, which adopts an UNet for progressive denoising in a learned latent space of images. Instead of training an independent network branch or new layers from scratch to process the new visual condition, we directly encode the reference image into the same latent space using the VAE. 
More specifically, we augment the first layer of SD UNet to simultaneously accept the noisy image latent and the reference latent as the inputs.
We then finetune the UNet to learn the natural blending of the visual references and the textual instruction to generate high-fidelity images following both conditions. 
Considering the large scales of the state-of-the-art diffusion models, tuning full models for diverse tasks can be prohibitive in terms of both parameter size and memory footprints. We resort to the low-rank adaptation of the SD models and develop a series of small-scale expert plug-ins, each containing only up to 28.55M parameters. 
Training objectives and auxiliary networks are customized for various applications, including English scene-text generation, multi-lingual scene-text generation, and logo-image generation.
We present comprehensive discussions on the sampling strategies and showcase the distinct impacts of both conditions in different denoising time steps, achieving an  English scene-text, multi-lingual scene-text, and logo generation accuracy of 61.73\%, 46.88\%, 44.07 with high prompt fidelity, outperforming the baselines.

The promising results on diverse applications shed light on a general way of customizing expert text-to-image models for particular subjects of interest.

\section{Related Works}
\label{sec:related_works}

\noindent\textbf{Diffusion models.} Diffusion models 
Surpassing the prior family of generative models such as GAN \cite{gan,wgan,dcgan,stylegan} VAE \cite{vae1,vae2,cvae,vqvae}, diffusion models \cite{sohl2015deep,ho_diffusion,song2020score} have demonstrated remarkable capabilities in generating images with both high quality and diversity, either in pixel space \cite{ho_diffusion,ddim,edm} or a learned latent space \cite{SD,unleashing,bld}. 
With the help of advanced pretrained language models \cite{clip,t5} and sampling techniques \cite{ddim,cfg}, text-to-image diffusion models \cite{SD,imagen,glide,chang2023muse,podell2023sdxl,ediffi,raphael} show unprecedented results on image generation following textual instruction. 
They generate high-resolution images by either operating in latent space \cite{SD,gu2022vector} or using cascaded models \cite{imagen,ediffi} to progressively scale up resolution.  
Motivated by the known issue of misspelling for almost all the public text-to-image diffusion models, several works have been proposed to improve text drawing by additional conditions such as masks \cite{textdiffuser} and glyph \cite{glyphdraw}. However, the model design customized for a particular language \cite{textdiffuser} prevents them from generalizing to general visual references. 
Recent research has been striving to improve the generation quality of particular subjects by model tuning. Specifically, model customization \cite{dreambooth,multi,elite,suti,instantbooth,blip,break_a_scene} focuses on a transfer learning approach that tunes model parameters to fix new concepts given examples. Textual inversion \cite{ti} learns a new concept by learning a new word vector. These methods fit well to particular subjects but usually fail to generalize to other similar ones. A line of research works \cite{controlnet,controlnet2,reco,gligen,make_a_scene} such as ControlNets \cite{controlnet} have been proposed to enhance the controllability of the generation using a new input condition.

\noindent\textbf{Visual text generation.} Despite the progress in generating images in high-quality images, the existing generative model has been noted to generate visually inaccurate images (i.e., misspellings). Several research studies have been proposed to mitigate this issue \cite{drawtext,imagen,deepfloyd,glyphdraw,textdiffuser}. One line of research has shown that the visual accuracy of the text renderings can be improved by deploying a language model of larger capacity \cite{deepfloyd,imagen,glyphdraw,glyphcontrol}. Imagen \cite{imagen} has demonstrated that encoding text prompts with the generic text encoder T5 \cite{t5} pretrained on text-only corpora can improve both the fidelity and image-text alignment \cite{imagen}. To evaluate the accuracy of the generated visual text images, several papers have constructed their own benchmark sets \cite{drawtext,glyphdraw,textdiffuser} using off-the-shelf OCR models \cite{swinsptter,maskspotter,units,deepsolo,synspotter}.

While some existing works rely on text renderings to improve English and multi-lingual generation \cite{textdiffuser,glyphcontrol,anytext}, or focus on personalized object generation \cite{dreambooth,blip}, our work is distinguished by introducing a single, unified framework that bridges these capabilities. By leveraging lightweight, task-specific expert plugins, our method achieves high fidelity in producing accurate multi-lingual text as well as complex brand logos, thereby validating its broad applicability.


\begin{figure*}[t]
\begin{center}
    \includegraphics[width=\textwidth]{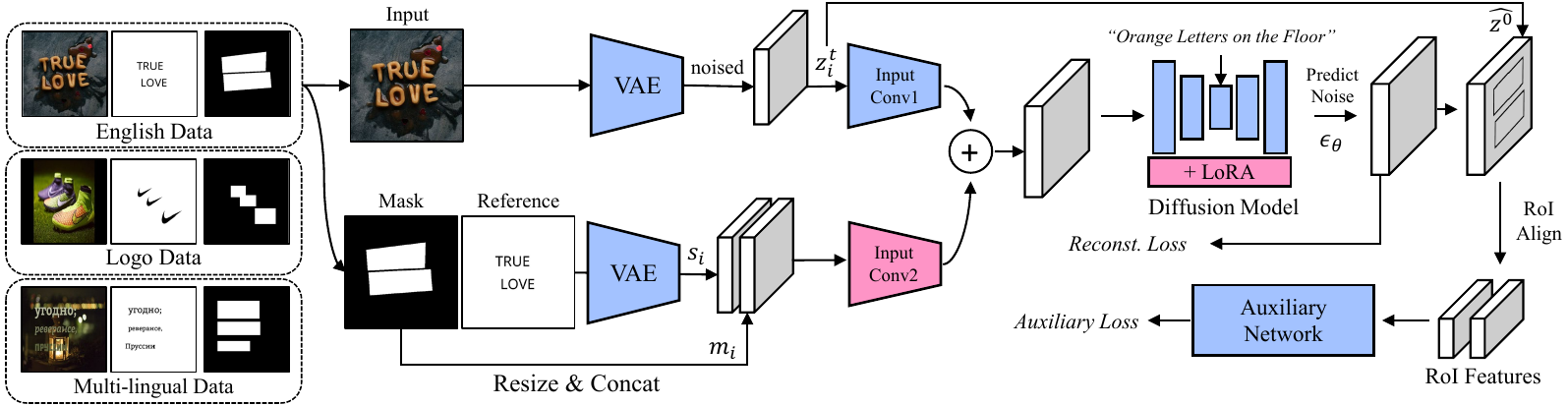}
\caption{\textbf{Overview of the model training.} Expert plugin modules are trained on each dataset. The input and reference conditions are encoded using a VAE, and the mask is resized and concatenated with the reference. An additional convolutional layer (Input Conv2) processes these concatenated features, and its output is combined with the existing input convolution layers. The regional features are obtained from denoised latents, and an auxiliary network receives them. The auxiliary network is trained online with each expert plugin to encourage accurate drawing of the target subjects. The Input Conv2 and the LoRA parameters are updated during training.}
\label{fig:overview}
\end{center}
\end{figure*}

\section{Method}    
The proposed \refdiffuser generates images conditioned on both a text prompt and a reference image. The image condition gives the model an uncurated visual reference for the generation targets, such as character shapes in scene-text image generation. We train the model to blend the reference concept naturally into the generated images without violating the text conditions. 
This reference condition empowers the model to produce contents that the original diffusion model fails to generate precisely, and can even help extend the generation to concepts not included in the language model vocabulary. 
Our proposed model is constructed on the foundation of the pre-trained Stable Diffusion model \cite{SD}. We employ the low-rank adaptation technique \cite{lora} to fine-tune the model for specific tasks, such as scene-text image generation, and develop a series of small-scale plug-ins, each seamlessly converting a Stable Diffusion model into an expert model for different applications with references. 

\newcommand{\modelinput}{\mathbf{x}}
\newcommand{\reflatent}{\mathbf{s}}
\newcommand{\imglatent}{\mathbf{z}}
\newcommand{\prompt}{\mathbf{c}}
\newcommand{\mask}{\mathbf{m}}
\subsection{Overview}
The proposed \refdiffuser builds upon the foundation of the pre-trained SD models \cite{SD} with a pre-trained CLIP \cite{clip} text encoder to encode text prompts, and is trained to incorporate additional image conditions. 
Training additional network branches to process the reference images introduces considerable cost \cite{controlnet}. Exploiting the fact that the references reside in the same image modality, we directly encode the reference images into the same latent space using the VAE \cite{vae} encoder of SD, and augment the input layer of the diffusion UNet \cite{unet}, expanding the inputs with the additional reference latent and a spatial mask indicating the location of the targets of interest in the reference image. 

Specifically, the input to the diffusion UNet is changed from the original $\imglatent_i^t\in\mathbb{R}^{c\times h\times w}$, which is the VAE encoded representation of the input image $i$ with noise injected at time step $t$, to $\modelinput_i^t$:
\begin{equation}
    \modelinput_{i}^t=\text{concat}(\imglatent_{i}^t, \reflatent_{i}, \mask_{i}),
    \label{eq:input_concat}
\end{equation}
where $\reflatent\in\mathbb{R}^{c\times h\times w}$ is a VAE encoded representation of the spatial reference image, and $t \sim [1, T]$ is the step within the $T$ total time steps. The location mask $\mask \in \mathbb{R}^{h \times w}$ in (\Cref{eq:input_concat}) is a binary array indicating the location and desired shape of the reference objects. The binary location mask $\mask$ is downsampled to the same spatial size as the latent.
The mask $\mask$ and reference latent $\reflatent$ remain constant across all time steps and are not injected with any noise in contrast to $\imglatent$.
With this input composition, we now have $\modelinput^t\in\mathbb{R}^{(2c+1)\times h\times w}$. 

To handle the additional $c+1$ channels, we introduce an extra input convolutional layer (\textit{Input Conv2}, \Cref{fig:overview}). Its output is added feature-wise to that of the original input convolution, and the combined feature map is then passed to the subsequent layers of the diffusion model for reconstruction. The reconstruction loss of diffusion training is defined as:
\begin{align}
    \mathcal{L}_{\text{diff}}=\sum_{i=1}^N||\epsilon_i-\epsilon_\theta(\modelinput^t_i, \prompt_i, t)||^2,
\end{align}
where $\epsilon_\theta$ represents the diffusion model parametrized by $\theta$, $N$ is the batch size, $\epsilon$ denotes the noise prediction target, and $\prompt_i$ is the CLIP embedding of the text prompt. 

Directly tuning $\theta$ introduces a high cost in terms of both parameter sizes and memory footprints. Therefore, we resort to the low-rank adaptation method \cite{lora} and develop expert plug-ins for different applications, each containing only a small number of parameters.

\subsection{Example Plugin 1: Text Image Generation}

In scene-text image generation, our goal is to produce high-quality images that not only align with the text prompt describing a scene but also accurately generate the target text. To achieve precise text generation, we provide the model with an additional reference image in which the desired characters are pre-rendered at the corresponding location in the scene, as illustrated in \Cref{fig:overview}. We use OCR detection results \cite{textdiffuser} to locate texts within the images, which can easily be obtained with the standard OCR detectors \cite{swinsptter}. Specifically, given the OCR labels, we generate the reference image by rendering the text to the corresponding region on a blank canvas. We also generate the binary location mask, where positive values indicate the precise size and shape of the desired text to draw in the image. With the reference image encoded using the SD VAE for $\reflatent$ and binary mask $\mask$ resized to the same spatial size as the latent, we construct the model input as in (\Cref{eq:input_concat}).

To ensure the spelling accuracy of the character sequence within the mask region, we facilitate the training with an online-trained lightweight text recognition network. We detail the network architecture of the recognition network in \Cref{app:detail_auxnet}. Given the noise estimation predicted by the diffusion network, we reconstruct the denoised latent at timestep $0$, denoted as $\hat{\imglatent}^0$, from the corrupted latent $\imglatent^{t}$ \cite{ho_diffusion}. We employ RoIAlign \cite{mask_rcnn} to extract text regions into a fixed size, and these pooled regions are fed into an online-trained text recognition network $\boldsymbol{\psi_{\text{recog}}}(\cdot)$ to compute the text recognition loss. The recognition loss is computed as,
\begin{align}
\small
    \textbf{r}_k&=\text{RoIAlign}(\hat{\imglatent}^0,\mathbf{B}_k) 
    \\
    \textbf{o}_k&=\psi_{\text{recog}}(\textbf{r}_k) \\
    \mathcal{L}_{\text{recog}}&=\frac{-1}{L}\sum_{j=1}^{L}y_{j}\log \textbf{o}_{k,j},
\end{align}
where $\textbf{B}_k$ is the bounding box label for $k$-th region in the image, $\textbf{o}_k$ is the network output. $L$ denotes the length of the $k$-th word and $j$ is an enumerator for each character in the word. Here, $\textbf{o}_{k,j}\in \mathbb{R}^{|C|}$ is the predicted probability over the character set at each position $j$, and $y_{j}\in\mathbb{R}^{|C|}$ is the corresponding one-hot label. We provide more details of the loss in \Cref{app:detail_auxnet}. Note that, unlike TextDiffuser, our method only requires word-level information, instead of character-level segmentation supervision \cite{textdiffuser}. This network can be discarded after training, and the final model consists of only the original SD UNet and a lightweight expert plug-in. The final learning objective of the diffusion network, $\mathcal{L}_{\epsilon}$, combines the reconstruction loss and the recognition loss through a weighted sum as:
\begin{equation}
\small
\mathcal{L_\epsilon}=\lambda_{\text{recog}}\mathcal{L_\text{recog}}+\lambda_{\text{diff}}\mathcal{L_\text{diff}}.
\end{equation}


\newcommand{\TSPw}{0.1\linewidth}
\newcommand{\TSPh}{0.05\linewidth}
\begin{figure*}
\centering
{\includegraphics[width=\linewidth]{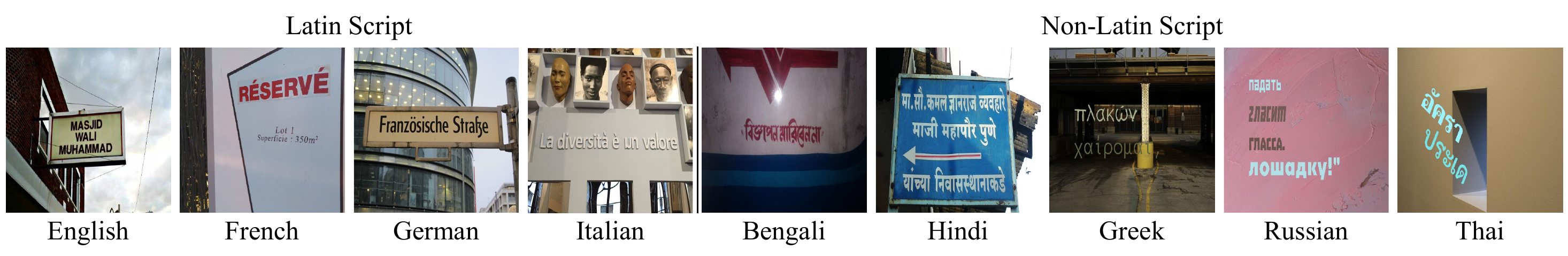}} 
\caption{\textbf{Example images from the MLT training dataset.} \textbf{Left:} Languages in Latin scripts. \textbf{Right:} Languages in Non-Latin script. We use synthetic images for Greek, Russian, and Thai languages.
}
\label{fig:mlt_examples}
\end{figure*}

\smallbreak
\noindent\textbf{Multi-Lingual Text Image Generation.} 
The flexibility of the conditions allows us to extend the generation to concepts beyond the vocabulary of the language model SD trained with. To show this, we develop an expert plug-in for multi-lingual text (MLT) image generation that covers multiple languages. 
Due to the difficulties in collecting large-scale datasets for MLT generation, we first pretrain the plugin module and recognition network on the large-scale English OCR dataset \cite{textdiffuser}, then proceed with the fine-tuning. The overall learning process is mostly the same as the English text image generation training, except that the size of the alphabet is expanded to cover all the target languages simultaneously. Due to insufficient images in the existing MLT OCR datasets, we augment training data with synthesized images by manually collecting text-free background images \cite{pexels}, fonts in different scripts, and text corpora of each language, then rendering text words in random regions inside the background images (e.g., \Cref{fig:mlt_examples}, Greek, Russian, Thai.). 
We then use a merged dataset with both real images from MLT OCR datasets \cite{icdar2019} and the synthesized image. Directly training the model with the mixed dataset compromises the quality of the generated images, as they can exhibit noticeable artifacts inherited from the synthetic images. 
To remedy this, we introduce an additional scaler $\alpha$ to scale the diffusion reconstruction loss:
\begin{equation}
\begin{aligned}
    \mathcal{\widetilde{L}}_{\text{diff}}=&\sum_{i\in\text{synth}}^N \alpha||\epsilon_i-\epsilon_\theta(\modelinput^t_i, \prompt_i, t)||^2  \\
    &+\sum_{i\notin\text{synth}}^N ||\epsilon_i-\epsilon_\theta(\modelinput^t_i, \prompt_i, t)||^2.
\end{aligned}
\end{equation}
We scale down $\alpha$ for the synthetic images to prevent the diffusion model from overfitting to the synthetic artifacts when learning to compose the scene and exploit the synthetic data mainly for improving the spelling correctness with the help of the recognition network. We observe that this simple loss scaling effectively improves the generated image quality. Loss terms related to the recognition network are identical to the ones used in the English scene-text generation.

\subsection{Example Plugin 2: Logo Image Generation} 
We further show that the proposed framework can be generalized to non-text objects and use logo image generation as an example to show the versatility of the proposed conditional image generation with reference guidance. 

\begin{figure}[htb!]
\includegraphics[width=\columnwidth]{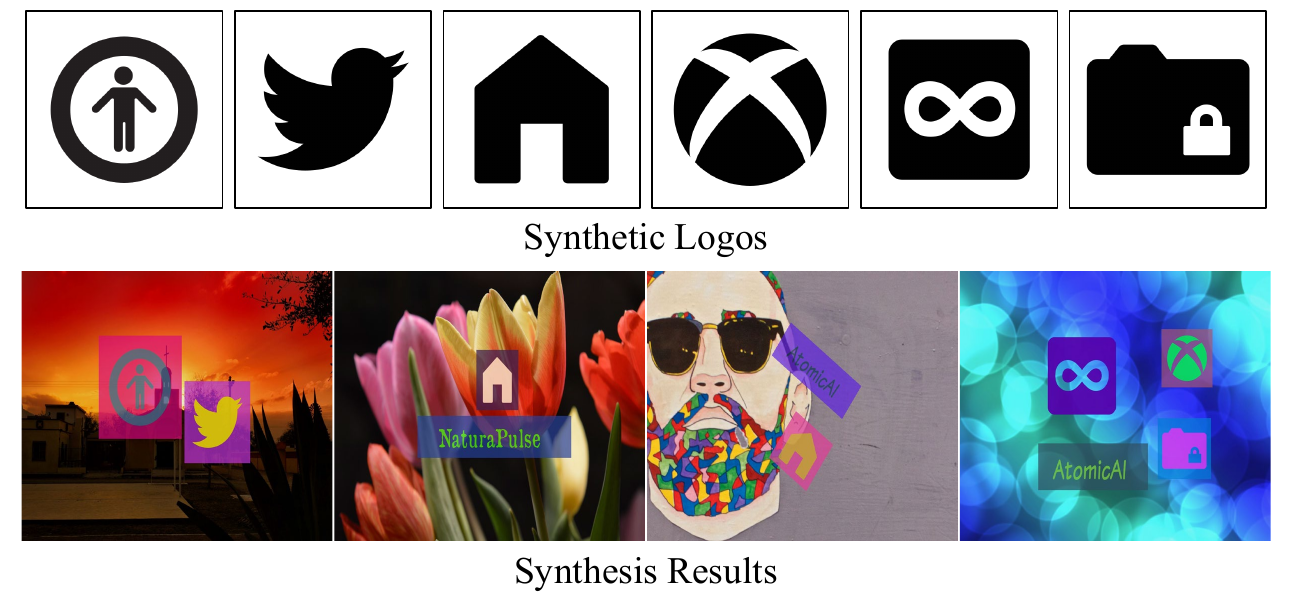}
\caption{\textbf{Example synthesis result of the log images.} \textbf{Top:} Example logo images that are embedded into background images. \textbf{Bottom:} The resulting synthesized images using the logos.}
\label{fig:synthesis_logo}
\end{figure}

For the reference images of logo image generation, we paste the standard logo we collected from the Internet onto a blank canvas and create the reference images. We visualize one example reference image in \Cref{fig:overview}. Similar to the MLT case, we synthesize logo images to augment the training set (\Cref{fig:synthesis_logo}), and we deploy an auxiliary network to classify logos. To compute the loss, we predict the denoised latent features at timestep $0$, $\hat{\imglatent}^0$, and apply the RoIAlign \cite{mask_rcnn} to extract the region $\textbf{r}_k=\text{RoIAlign}(\hat{\textbf{z}}^0,\textbf{B}_k)$. The RoI features are then provided to the auxiliary network $\psi_{\text{logo}}(\cdot)$ to do the classification:
\begin{equation}
    \mathcal{L}_{\text{logo}}=\frac{-1}{K}\sum_{k=1}^{K}y_{k}\log \psi_{\text{logo}}(\textbf{r}_k),
\end{equation}
where $K$ denotes the total number of RoI instances in the batch enumerated by $k$, $y_{k}\in\mathbb{R}^{|M|}$ is the one-hot labels for $|M|$ logos we use for training. The overall learning objective can be formulated as follows:
\begin{equation}
\small
\mathcal{L}=\lambda_{\text{logo}}\mathcal{L}_{\text{logo}}+\lambda_{\text{diff}}\mathcal{\widetilde{L}}_{\text{diff}}.
\end{equation}
Although we train the model with a closed set of $|M|$ logos, the model can generalize to the ones that are unseen during training. We show visual examples to validate this in \Cref{fig:unseen}, and \Cref{fig:app_unseen_logos} in the Appendix.

\subsection{Scheduled Classifier-Free Guidance}
\label{sec:inference}
\noindent{\textbf{Classifier-free guidance}} \cite{cfg} has become the standard technique for sampling with conditional diffusion models. It substantially improves the generated image quality and alignment with the condition prompt by extrapolating the predicted $\epsilon$ toward the direction of the condition:
\begin{equation}
\small
    \hat{\epsilon_\theta}^t := \epsilon_\theta(\imglatent^t, \prompt, t) + \omega (\epsilon_\theta(\imglatent^t, \prompt, t) - \epsilon_\theta(\imglatent^t, \varnothing, t)),
\end{equation}
where $\epsilon_\theta(\imglatent^t, \varnothing, t)$ denotes unconditional $\epsilon$ prediction with an empty text prompt.

\noindent{\textbf{Augmented Classifier-Free Guidance}.} With the additional reference condition included, we opt to reformulate the CFG as below to enable more detailed guidance of each condition,
\begin{align}
\small
\begin{split}
    \hat{\epsilon}^t & = \epsilon_\theta(\text{concat}(\imglatent^t, \varnothing, \varnothing),  \varnothing, t) \\
    & + \omega_{\text{prompt}} \Big[\epsilon_\theta(\text{concat}(\imglatent^t, \varnothing, \varnothing), \prompt, t) - \epsilon_\theta(\text{concat}(\imglatent^t, \varnothing, \varnothing), \varnothing, t)\Big] \\
    &+ \omega_{\text{ref}}\Big[\epsilon_\theta(\text{concat}(\imglatent^t,\textbf{s},\textbf{m}), \varnothing, t) - \epsilon_\theta(\text{concat}(\imglatent^t, \varnothing, \varnothing), \varnothing, t)\Big] \\
    &+ \omega_{\text{all}}\Big[\epsilon(\text{concat}(\imglatent^t,\textbf{s},\textbf{m}), \prompt, t) - \epsilon(\text{concat}(\imglatent^t, \varnothing, \varnothing), \varnothing, t)\Big],
\end{split}
\label{eqn:cfg_schedule}
\end{align}
where the reference condition is set to be full zero in $\text{concat}(\imglatent^t, \varnothing, \varnothing)$ to omit the reference condition. To enable the updated classifier-free guidance, we randomly drop the text condition $\prompt$ and the reference condition $\{\reflatent, \mask\}$ at a 10\% chance during training. We provide additional details of the augmented CFG in \Cref{app:detail_aug_cfg}.

\noindent{\textbf{CFG-scale scheduling.}} Standard classifier-free guidance adopts consistent scale $\omega$ across all sampling steps. We empirically observe that when an additional condition is introduced, dynamically adjusting the scale for each condition across sampling steps can further improve generated results.
Specifically, we observe that setting a higher guidance scale $\omega_\text{ref}$ for reference conditions at the earlier sampling steps benefits the accurate visual text rendition, as it helps to establish the overall layout of the text elements. Hence, we dynamically adjust the guidance scales as,
\begin{align}
\begin{split}
    \omega^t_\text{ref} &= \gamma\frac{t}{T}^{\rho_\text{speed}}, \\
    \omega^t_\text{prompt} &= \gamma(1-\frac{t}{T}^{\rho_\text{speed}}).
\end{split}
\end{align}
We set $\gamma$ and $\omega_\text{all}$ as constant values, and $\rho_\text{speed}$ is also a constant that controls the speed of increase or decrease in the guidance scale.
\section{Experiments}

\begin{figure}[t]
    \centering
    \includegraphics[width =\linewidth]{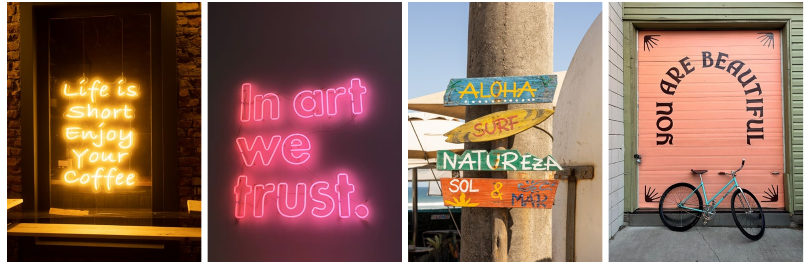}
\caption{\textbf{Example images in the AText dataset we collect.}}
\label{fig:pexel}
\end{figure}

\begin{table*}[t]
\centering
\begin{minipage}{0.95\linewidth}
    \begin{minipage}{0.32\linewidth}
        \centering
        \small
        \begin{tabular}{c|cc}
            \toprule
            \textbf{Method} & \textbf{Acc}$\uparrow$& \textbf{CLIP}$\uparrow$ \\
            \midrule
            SD                          & 0.00          & 0.2902         \\
            TextDiffuser                & 0.00          & 0.3045         \\
            ControlNet                  & 15.00          & \textbf{0.3054}\\
            Ours$^{\text{\textendash}}$ & \textbf{22.50} & 0.3037         \\
            \bottomrule
        \end{tabular}
        \caption{\textbf{Evaluation results of zero-shot Russian generation}. Ours$^{\text{\textendash}}$ denotes the model trained \emph{without} seeing any Russian images during training. The accuracy is denoted in [\%].} 
        \label{tab:zeroshot}
    \end{minipage}
    \hspace{3mm}
    \begin{minipage}{0.32\linewidth}
    \vspace{-15mm}
    \small
        \begin{center}
            \begin{tabular}{c|cc}
                \toprule
                \textbf{Method} & \textbf{Acc}$\uparrow$ & \textbf{F-1}$\uparrow$ \\
                \midrule
                w/o Reference   &2.00   & 27.93\\
                w/ Reference  &\textbf{46.13} & \textbf{74.58} \\
                \bottomrule
            \end{tabular}
            \caption{\textbf{Ablation studies on the impact of reference image.} Results are denoted in [\%].}
            \label{tab:abl_reference}
        \end{center}
    \end{minipage}
    \hspace{3mm}
    \begin{minipage}{0.3\linewidth}
    \vspace{-14mm}
    \small
        \centering
        \centering
        \begin{tabular}{c|cc}
            \toprule
            \textbf{Method} & \textbf{\# Params (M)}$\downarrow$ \\
            \midrule
            SD            & 859.52 \\
            TextDiffuser  & 876.86 \\
            ControlNet    & 1220.80 \\
            Ours          &\textbf{28.55} \\
            \bottomrule
        \end{tabular}
    \caption{\textbf{Comparison results of the number of trainable parameters.}} 
    \label{tab:cost}
    \end{minipage}
\end{minipage}
\end{table*}

\subsection{Datasets}
We explain the dataset that we use for training. Additional details are available in \Cref{app:dataset,app:benchmark_details}.

\noindent{\textbf{English.}} For training English plug-in models, we first train the model with MARIO-10M \cite{textdiffuser} and further fine-tune the model by including samples from LAION-Aesthetic \cite{laion} and additional aesthetic text images (AText) that we collect (\Cref{fig:pexel}). We use MARIO-Eval \cite{textdiffuser} benchmark for evaluation.

\noindent{\textbf{Multi-Lingual Text.}} For the MLT plug-in, we finetune the English plug-in model trained with MARIO-10M using the merged set of real images from ICDAR2019 \cite{icdar2019}, and synthetic images that we generated. Due to the lack of pre-existing benchmarks for MLT image generation, we developed a new benchmark for model evaluation on nine different languages, having 2,500 prompts in total. Details of the benchmark are available in \Cref{app:benchmark_details}.

\noindent{\textbf{Logos.}} Similar to the MLT plug-in, we also finetune the English plug-in model trained with MARIO-10M using the merged set of real images from the merged set of FlickrLogos-32 \cite{flickr27} and Logos in the wild (LITW) \cite{litw}, and the synthetic images that we generated. For evaluation

\subsection{Evaluation Metrics}
We use \textbf{Fréchet inception distance (FID)} to assess the image quality, and we evaluate the semantic alignment by computing the \textbf{CLIP Score} between the text prompts and the generated images and measuring the aesthetic scores of the generated images obtained from a pretrained model \cite{laion}.

\noindent\textbf{Text images.} To measure the accuracy of the generated text, we use \textbf{Accuracy} and \textbf{F1-score}. 
We apply different OCR Engines for English generation tasks and MLT image generation tasks. We use the Microsoft Read API for English, and the Google Cloud Vision API for MLT image generation tasks. 
We consider a detected text correct if it is exactly matched with a keyword, and the F-1 score is defined as the harmonic mean of precision and recall.

\noindent\textbf{Logo images.}  To measure the accuracy of the logo image generation, we train a Mask-RCNN \cite{mask_rcnn} on the merged set of FlickrLogos-32 \cite{flickr32} and LITW dataset \cite{litw}. We report the Accuracy and F1-score of the logo detection model applied to an image set generated by models.

\begin{table}[t!]
\begin{center}
\footnotesize
  \begin{tabular}{ccccc}
    \toprule
    \textbf{Method}    & \textbf{FID}$\downarrow$   & \textbf{CLIP}$\uparrow$ &\textbf{ Acc}$\uparrow$  &\textbf{F-1}$\uparrow$.\\
    \midrule
    SD & 51.29   & 0.3015    & 0.03 & 2.14 \\
    ControlNet & 51.48   & 0.3424     & 23.90  & 58.65 \\
    DeepFloyd& \textbf{34.90}   &0.3267    & 2.62 & 17.62 \\
    TextDiffuser  & 38.75   & 0.3436    & 56.09  & 78.24 \\
    SD3  & 37.21   & 0.3424          & 9.62  & 49.55 \\
    SDXL  & 58.54   & 0.3242         & 2.16  & 20.33 \\
    Flux  & 41.25   & 0.3198         & 54.98  & 52.35 \\
    GlyphControl  & 61.10   & 0.3411 & 27.04  & 64.00 \\
    AnyText  & 52.39   & 0.3426      & 18.03  & 60.74 \\
    \midrule
    Ours  & 38.59   & \textbf{0.3454}    & 58.26 &  79.15 \\
    Ours$^+$  & 42.19   & 0.3434    & \textbf{61.73} &  \textbf{80.08} \\
    \bottomrule
  \end{tabular}
  \caption{\textbf{Comparison results of English generation on MARIO-Eval benchmark}. Acc and F-1 denote OCR Accuracy and F1-score, respectively. $+$ denotes the results with scheduled classifier guidance. Accuracy and F-1 are denoted in [\%].}
\label{tab:english_eval}
\end{center}
\end{table}

\subsection{Implementation Details.} 
\label{sec:implement_detail}
We first fine-tune the pre-trained Stable Diffusion (SD) V2.1 \cite{SD} on the MARIO-10M dataset \cite{textdiffuser}. For English image generation, we further train the model using the AText dataset we collect. For MLT and Logo image generation, we train the model with merged sets of synthetic images and real images. We set batch size=10 for each of the 8 GPUs. We use AdamW \cite{adamw} with a learning rate of 1e-4 to tune the parameter. For MLT image generation, we use a character set containing 847 characters from the nine languages. We use a rank of 32 for all LoRA. We use 100 steps of DDIM \cite{ddim} sampling. For all the experiments, we set the recognition loss weight $\lambda$ as 0.025.

\begin{table*}[t!]
    \centering
    \footnotesize
        \begin{tabular}{c|cccc|cccc}
        \toprule
        \multirow{2}{*}{\textbf{Method}}   &\multicolumn{4}{c|}{\textbf{Latin}}&\multicolumn{4}{c}{\textbf{Non-latin}}\\ \cline{2-9}
          &\textbf{FID}$\downarrow$   & \textbf{CLIP}$\uparrow$ &\textbf{Acc}$\uparrow$  &\textbf{F-1}$\uparrow$&\textbf{FID}$\downarrow$   & \textbf{CLIP}$\uparrow$ &\textbf{Acc}$\uparrow$  &\textbf{F-1}$\uparrow$ \\
        \midrule
        SD          & 114.34&0.3032&0.38&1.87   
                    &113.07&0.3067& 0.00 & 0.00  \\ 
        GlyphControl& 115.51 & 0.2922 &58.25   & 68.73 
                    &130.44  & 0.2879 & 0.00   & 0.00 \\
        AnyText     & 115.51 & 0.3100 &  10.38 & 48.87 
                    &130.44  & 0.3155 & 4.50   & 17.42 \\
        SD3         &134.31  & 0.2983 & 25.06  & 46.13 
                    &133.24  & 0.3090 & 0.11   & 0.11 \\
        SDXL        &126.49  & 0.3041   & 16.18  & 13.90 
                    &132.13  & 0.3101   & 0.00   & 0.00 \\
        ControlNet  &140.03   &0.3010&10.63&43.03  
                    &136.54   &0.3038 & 12.00 & 40.09 \\ 
        \midrule
        Ours        &  117.20&0.2952  &56.38  & 76.68 
                    &116.79  & 0.3059 &23.80 & 38.96 \\
        Ours$^+$    &  117.96& 0.2803  &64.56  & 79.09 
                    &117.09  & 0.2940 &29.20 & 41.78 \\
        \bottomrule
    \end{tabular}
    \caption{\textbf{Evaluation results of the MLT image generation.} \textbf{Latin languages}: English, German, French, and Italian. \textbf{Non-Latin languages}: Bengali, Hindi, Greek, Russian, and Thai. Result denoted by $+$ applies CFG scheduling with $\rho_\text{speed}=0.2$, $\gamma=3.5$ and $\omega_\text{all}=4$. Accuracy and F-1 are denoted in [\%].}
    \label{tab:mlt_eval}
    \vspace{-1mm}
\end{table*}

\subsection{Ablation Study}
In this section, we present ablation studies to gain deeper insights into the proposed method. We provide additional ablation studies in \Cref{app:ablation}.

\noindent{\textbf{Impact of Reference Image.}} 
We examine the influence of reference image guidance. Notably, we observe a huge gap in the performance between the model trained to be conditioned with additional reference guidance and the model without it (\Cref{tab:abl_reference}). This observation indicates the significant role that visual references play in both proper conditioning and providing valuable information about the targets.

\noindent{\textbf{Zero-shot Text Image Generation.}} As reference guidance enables our model to handle conditions beyond the vocabularies encountered during training, we explore its zero-shot capability. We evaluate the zero-shot MLT results for Russians by excluding all Russian text images during training \Cref{tab:zeroshot}. Notably, our zero-shot model outperforms the baselines in terms of OCR accuracy, demonstrating its effectiveness.

\noindent{\textbf{Efficiency Analysis.}} We report the parameter count with comparison with baseline approaches in \Cref{tab:cost}. Notably, our method requires significantly fewer parameters, demonstrating that it is both effective and resource-efficient, enabling efficient integration into existing pipelines.

\subsection{Quantitative Results}
We provide comparison results with SD \cite{SD}, SD3 \cite{sd3}, SDXL \cite{podell2023sdxl}, Flux \cite{flux}, GlyphControl \cite{glyphcontrol}, AnyText \cite{anytext}, ControlNet \cite{controlnet}, DeepFloyd \cite{deepfloyd}, and TextDiffuser \cite{textdiffuser}. For ControlNet, we use the model trained to be conditioned on the Canny Edge map, and we provide Canny Edge maps of rendered text images or logo images as input conditions during inference. For log generation, we also compare with BLIP-Diffusion \cite{blipdiff}, IP-Adapter \cite{ipadapter}, and MS-Diffusion \cite{msdiff}. We present results for our method both with and without the scheduled CFG guidance described in \Cref{sec:inference}.

\noindent\textbf{English Images.} 
For the English image generation task, our method achieves the best performance on all the OCR-related metrics, including OCR Accuracy and F-1 Score (\Cref{tab:english_eval}). Additionally, our method also achieves the best CLIP score, demonstrating that our model is not only capable of generating accurate images but also best aligns with the text prompts. Besides, our method achieves comparable results of FID with DeepFloyd, which achieves lower accuracy than ours by a large margin. When the proposed CFG scheduling scheme is applied, we obtain further improvement in accuracy, with a decrease in FID performance. We provide additional analysis on CFG scheduling at \Cref{app:ablation}.

\noindent\textbf{Multi-Lingual Images.} For multi-lingual text (MLT) image generation, we report results in the Latin language and the non-Latin language sets (\Cref{tab:mlt_eval}). There are five different languages in \textit{Latin languages}: English, German, French, and Italian. There are four different languages in \textit{Non-Latin languages}: Bengali, Hindi, Greek, and Russian. For all language sets, we achieve the best accuracy result. Similar to English results, we obtain improvements in accuracy when CFG scheduling is applied. We report the detailed per-language results in \Cref{app:abl_lang} due to the space limit.

\noindent\textbf{Logo Images.} 
Our approach also attains notable results in producing logo images. Our model ranks second in the FID measurement, while achieving the best results on CLIP score, Accuracy, and F-1 score (\Cref{tab:logo_comparison}). Similar to English and MLT generation, the model achieves improvement in accuracy when CFG scheduling is applied. To further demonstrate the generalization of the trained plug-in for logo image generation, we fed the network with novel logos and icons that are not included in the dataset. Apart from offering natural and faithful blending of the logos to the described scenes, we further demonstrate in \Cref{fig:unseen} that our method generalizes to novel logos and icons unseen during model training.


\begin{table}[t!]
    \centering
    \footnotesize
    \begin{tabular}{ccccc}
    \toprule
    \textbf{Method}&\textbf{FID}$\downarrow$ & \textbf{CLIP}$\uparrow$& \textbf{Acc}$\uparrow$&\textbf{F-1}$\uparrow$\\
    \midrule
    SD        &\textbf{74.40}& 0.3469   & 10.33   & 13.71\\
    ControlNet    & 110.77   & 0.3553   & 34.20   & 39.36\\
    TextDiffuser  & 88.97    & 0.3183   & 9.13 & 10.06 \\
    BLIP-Diffusion& 104.18   & 0.3418   & 12.20 & 18.27 \\
    IP-Adapter    & 117.72   & 0.3509   & 22.47 & 25.44 \\
    MS-Diffusion  & 83.27    & 0.3769   & 34.73 & 41.23 \\
    \midrule
    Ours          &88.49 & 0.3759         & 42.87        & 48.86\\
    Ours$^+$      &89.49 & \textbf{0.3770} &\textbf{44.07}& \textbf{48.91}\\
    \bottomrule
    \end{tabular}
    \caption{\textbf{Evaluation results for logo image generation.} Result denoted by $+$ applies CFG scheduling with $\rho_\text{speed}=0.2$, $\gamma=3.5$ and $\omega_\text{all}=4$ for CFG scheduling. Accuracy and F-1 are denoted in [\%].}
    \label{tab:logo_comparison}
    
\end{table}

\begin{figure*}[h!]
\centering
 \includegraphics[width=0.92\linewidth]{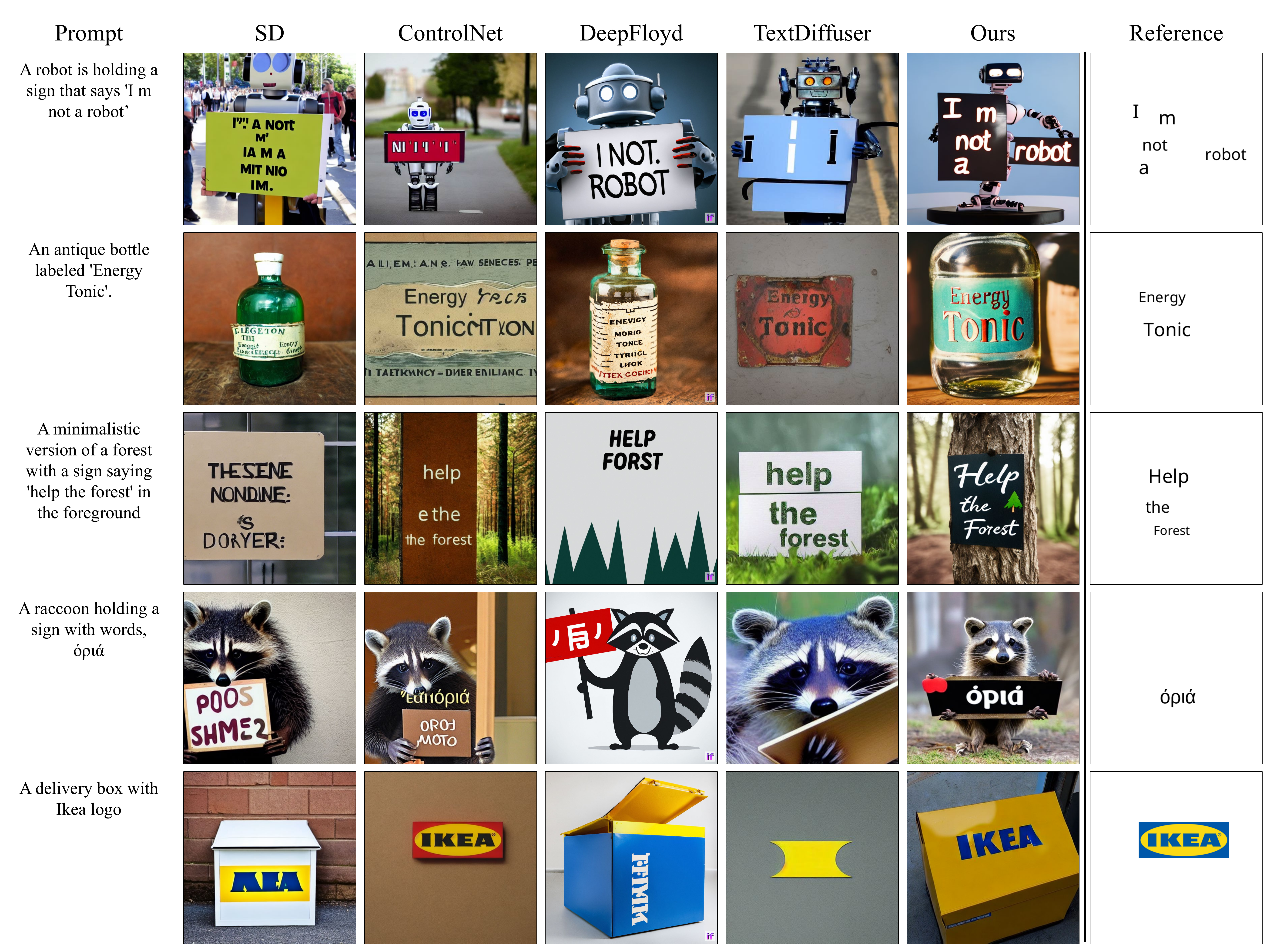}
 \caption{\textbf{Side-by-side qualitative comparison with baselines.}}
 \label{fig:sidebyside}
\end{figure*}
\subsection{Qualitative Results}
\begin{figure}[t!]
\centering
 \includegraphics[width=1\linewidth]{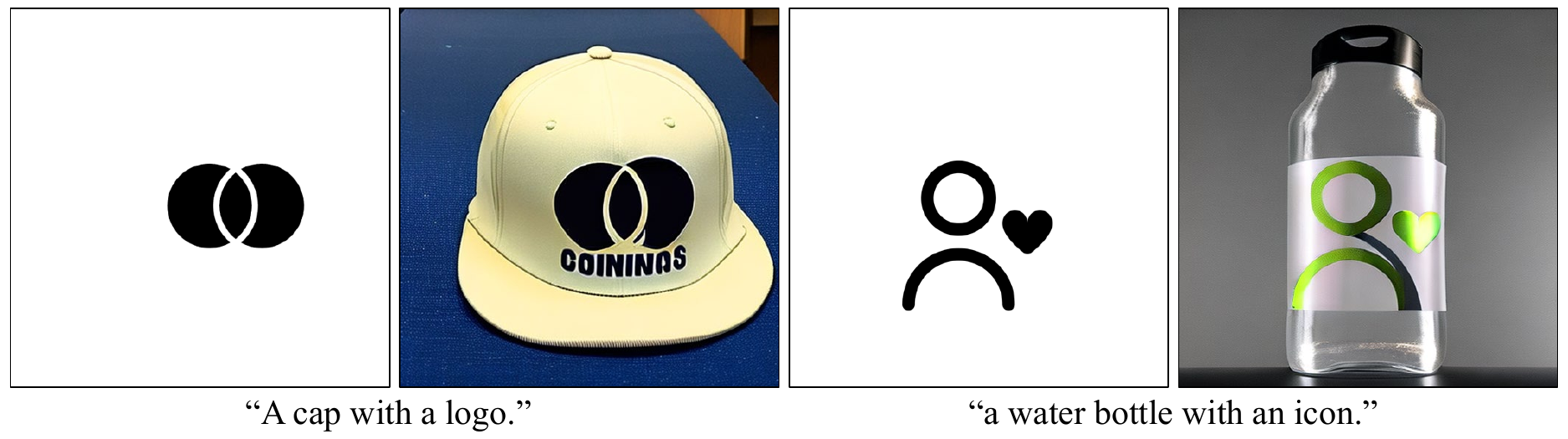}
 \caption{\textbf{Unseen logo image generation results.} The logo images in the reference are not included in the training logo set and hence, are unseen during the model training. More visualization results of generated unseen logo images are provided in \Cref{app:results}.}
 \label{fig:unseen}
\end{figure}

\begin{figure}[t!]
\centering
 \includegraphics[width=1\linewidth]{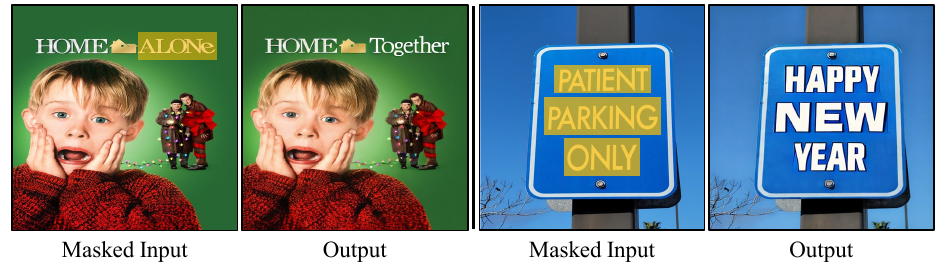}
 \caption{\textbf{Text editing results.} Regions denoted in yellow are masked and edited.}
 \label{fig:inpaint}
\end{figure}

We present side-by-side results in \Cref{fig:sidebyside} for direct comparisons of the image quality of different methods. Our method achieves both faithful blending of target subjects to the scene and strong correspondence to the text prompt in all experiments. 
We also provide logo image generation results when unseen logo images are given (\Cref{fig:unseen}), which further validates the effectiveness of the proposed method and the model's ability to generalize across novel instances.
Moreover, our method is highly flexible and can be easily extended to text editing tasks (\Cref{fig:inpaint}). Following \cite{textdiffuser}, we augment the channel of the input latent by concatenating the encoded latent of the masked image. We provide more details and additional qualitative results in \Cref{app:results}.
%

\section{Conclusion}
In this paper, we introduced \textsc{RefDiffuser}, a text-to-image diffusion model based on a visual reference guide. We expanded a pretrained SD model to accept an additional reference image as input, which provides the model with visual guidance to the appearance of the generation target, allowing for the precise generation of concepts even beyond the text-encoder vocabulary. 
With lightweight expert plugins efficiently tuned by applying low-rank adaptation, and training methods adapted for each task, we demonstrated expert plugins for applications including English scene-text generation, multi-lingual scene-text generation, and logo-image generation.  
Experimental results validated the superiority of the proposed method, as our model achieves superior results on every task. This research shed light on a general framework for providing additional visual references to text-to-image models for precise generation.

\clearpage
{
    \small
    \bibliographystyle{ieeenat_fullname}
    \bibliography{main}
}

\clearpage
\appendix

\setcounter{page}{1}
\setcounter{table}{0}
\setcounter{equation}{0}
\setcounter{figure}{0}
\renewcommand{\thetable}{\Alph{table}}
\renewcommand{\thefigure}{\Alph{figure}}

\maketitlesupplementary

\section{Additional Technical Details}
\label{app:details}

\subsection{Auxiliary Network Architecture} 
\label{app:detail_auxnet}
We provide details of the auxiliary recognition network architecture. The auxiliary network is composed of the encoder and the recognition head. The encoders for the text image generation and logo image generation share the same encoder structure. The shared encoder network consists of multiple encoding blocks, each sharing the structure of the blocks in the VAE network of the diffusion model. More specifically, each encoder block comprises a GroupNorm layer, two Residual blocks, and another GroupNorm layer, where the number of groups is set to 16. We use four encoding blocks for both text recognition and log recognition networks. Except for the last encoder block, the spatial dimension is down-sampled by the factor of 2 using the convolution layer with stride=2.

\paragraph{Text recognition head.}
The text recognition head is composed of four cross-attention layers, where the cross-attention is computed between the sequence of input image tokens and the sequence of positional embeddings corresponding to the characters of the output text word. The cross-attention we compute here can be formulated as:
\begin{align}
    \text{Attention}(\mathbf{r},\Tilde{\imglatent},\Tilde{\imglatent})=\text{   softmax}(\frac{\mathbf{r}\Tilde{\imglatent}^T}{\sqrt{d_k}})\Tilde{\imglatent},
\end{align}
where $\mathbf{r}$ is the sequence of positional embeddings corresponding to $i^{\text{th}}$ character of the word, and $\Tilde{\imglatent}$ is the encoding from the auxiliary encoder.

\paragraph{Logo recognition head.}
The recognition head for the auxiliary logo recognition network is composed of a single fully connected layer, where the output dimension corresponds to the size of the logo set.

\paragraph{Text Recognition Loss.}
The text recognition loss that we use is defined as below,
\begin{align}
\small
    \textbf{r}_k&=\text{RoIAlign}(\hat{\imglatent}^0,\mathbf{B}_k) 
    \\
    \textbf{o}_k&=\psi_{\text{recog}}(\textbf{r}_k) \\
    \mathcal{L}_{\text{recog}}&=\frac{-1}{L}\sum_{j=1}^{L}y_{j}\log \textbf{o}_{k,j},
\end{align}
where $\textbf{B}_k$ is the bounding box label for $k$-th region in the image, $\textbf{o}_k$ is the network output, and $L$ denotes the length of the word in the $k$-th region. $j$ is an enumerator for each character in the word. We predict the probability distribution over the character set at each character position $j$ denoted as $\textbf{o}_k,j\in \mathbb{R}^{|C|}$. For each position, we apply cross-entropy loss between the one-hot label $y_{j}\in\mathbb{R}^{|C|}$. For MLT loss, we use the same loss design, but the output dimension is extended to 847 to cover all the characters of the target languages.


\subsection{Additional Details of Augmented CFG}
\label{app:detail_aug_cfg}

\paragraph{Classifier-free guidance \cite{cfg}} considers sharpened posterior distribution $P_\theta(\imglatent|\textbf{c})\propto P_\theta(\imglatent)P_\theta(\textbf{c}|\imglatent)^\omega$. Using Bayes' rule for some timestep $t$,
\begin{align}
\small
\begin{split}
    \nabla_\imglatent \log P_\theta(\imglatent_t|\textbf{c})&=\nabla_{\imglatent_t}\log P_\theta(\imglatent_t)\\
    &+\omega(\nabla_{\imglatent_t} \log P_\theta(\imglatent_t|\textbf{c})-\nabla_{\imglatent_t}\log P_\theta(\imglatent_t)).
\end{split}
\end{align}
Since the score function is parameterized with $\epsilon_\theta$, we have
\begin{equation}
\small
    \hat{\epsilon_\theta}^t := \epsilon_\theta(\imglatent^t, \prompt, t) + \omega (\epsilon_\theta(\imglatent^t, \prompt, t) - \epsilon_\theta(\imglatent^t, \varnothing, t)),
\end{equation}
where $\epsilon_\theta(\imglatent^t, \varnothing, t)$ denotes unconditional $\epsilon$ prediction with an empty text prompt.

Similarly we model the posterior distribution that can be represented as,
\begin{align}
\small
\begin{split}
P_\theta(\imglatent|\textbf{s},\textbf{m},\textbf{c}) &\propto P_\theta(\imglatent)P_\theta(\textbf{s},\textbf{m},\textbf{c}|\imglatent)^{\omega_\text{all}} \\
&\quad \cdot P_\theta(\imglatent)P_\theta(\textbf{s},\textbf{m}|\imglatent)^{\omega_\text{prompt}} P_\theta(\textbf{c}|\imglatent)^{\omega_\text{ref}}.    
\end{split}
\end{align}
Similarly, for some timestep $t$,
\begin{align}
\footnotesize
\begin{split}
    \nabla_\imglatent \log P_\theta(\imglatent_t|\textbf{s},\textbf{m},\textbf{c}) 
    &= \nabla_{\imglatent_t}\log P(\imglatent_t) \\
    &\quad + \omega_\text{all}\left(\nabla_{\imglatent_t} \log P_\theta(\imglatent_t|\textbf{s},\textbf{m},\textbf{c}) - \nabla_{\imglatent_t}\log P_\theta(\imglatent_t)\right) \\
    &\quad + \omega_\text{ref}\left(\nabla_{\imglatent_t} \log P_\theta(\imglatent_t|\textbf{s},\textbf{m},\varnothing) - \nabla_{\imglatent_t}\log P_\theta(\imglatent_t)\right) \\
    &\quad + \omega_\text{prompt}\left(\nabla_{\imglatent_t} \log P_\theta(\imglatent_t|\varnothing,\varnothing,\textbf{c}) - \nabla_{\imglatent_t}\log P_\theta(\imglatent_t)\right).
\end{split}
\end{align}

From this, we derive,
\begin{align}
\label{eq:cfg_schedule_app}
\small
\begin{split}
    \hat{\epsilon}^t & = \underbrace{\epsilon_\theta(\text{concat}(\imglatent^t, \varnothing, \varnothing),  \varnothing, t)}_\text{Unconditional} \\
    & \underbrace{+ \omega_{\text{prompt}} \Big[\epsilon_\theta(\text{concat}(\imglatent^t, \varnothing, \varnothing), \prompt, t) - \epsilon_\theta(\text{concat}(\imglatent^t, \varnothing, \varnothing), \varnothing, t)\Big]}_\text{Caption Guidance} \\
    &+ \underbrace{\omega_{\text{ref}}\Big[\epsilon_\theta(\text{concat}(\imglatent^t,\textbf{s},\textbf{m}), \varnothing, t) - \epsilon_\theta(\text{concat}(\imglatent^t, \varnothing, \varnothing), \varnothing, t)\Big]}_\text{Reference Guidance} \\
    &+ \underbrace{\omega_{\text{all}}\Big[\epsilon(\text{concat}(\imglatent^t,\textbf{s},\textbf{m}), \prompt, t) - \epsilon(\text{concat}(\imglatent^t, \varnothing, \varnothing), \varnothing, t)\Big]}_\text{All Guidance},
\end{split}
\end{align}

\subsection{Additional Details of Datasets}
\label{app:dataset}

\paragraph{MARIO-10M \cite{textdiffuser}} is a compilation of data from various publicly accessible sources, including the LAION-400M \cite{laion400m}, TMDB \cite{tmdb}, and Open Library datasets. The images within this dataset are filtered by a text detector and annotated by an OCR model. A training set containing 10 million images with English-only annotations are utilized for training.

\paragraph{ICDAR2019 \cite{icdar2019}}
stands as a multi-lingual text (MLT) dataset initially curated for text detection and recognition. This dataset contains 10,000 training images and provides coordinates of each text word with corresponding character labels. We use a subset with six languages, including English, French, German, and Italian to train our MLT image generation from this dataset.

\paragraph{FlickLogos-32 \cite{flickr32}} contains 8,240 images with 32 logo brands. This dataset was originally collected for the logo retrieval and logo detection/recognition tasks. 
We use a subset of 2,240 images that contain at least one logo as the training set for the logo image generation task.

\paragraph{Logos in the wild (LITW) \cite{litw}} comprises 11,054 images obtained from the Google image search engine, featuring 871 unique brands.  We filter the dataset by excluding images with logos smaller than 45 pixels high and images with no logo. After filtering, a total of 4,206 images are obtained for training.

\paragraph{LAION-Aesthetics \cite{laion_aes}.}
LAION-5B \cite{laion} is a large-scale dataset of image-caption pairs. LAION-Aesthetics is the subset of LAION-5B, filtered by a trained aesthetic score predictor, and contains only high aesthetic score (above 6.5) images. This dataset does not provide any annotations of text or logo, and is included in our training to prevent overfitting and catastrophic forgetting.

\paragraph{Aesthetic text dataset (AText).}
We collect an additional dataset for English text generation with high-aesthetic and high-artistic images from the Internet \cite{pexels}. We measure the aesthetic scores of the images using a pretrained model from \cite{laion} and exclude those with low aesthetic scores.  We run an OCR model to detect text regions and annotate the text within the regions. Example images are shown in \Cref{fig:pexel}.

\paragraph{Synthetic logo/text dataset.}
Due to the limited numbers of MLT and logo training samples from public datasets, we augment the training set with synthesis. We use high-resolution images from public datasets as the background and render the target text or logo. For the multi-lingual experiment, we use an E-book text corpus \cite{corpus} and translate the text into the target languages of interest. For the logo image experiment, we use publicly available icon images \cite{icons} as the target objects to be rendered on the background images (\Cref{fig:synthesis_logo}). In total, we obtain 23,892 synthetic logo images and 184,912 synthetic MLT images.

\subsection{Additional Details of Benchmarks}
\label{app:benchmark_details}
\paragraph{Benchmark for English generation.}
We adopt the MARIO-Eval \cite{textdiffuser} benchmark to assess the model performance in generating English scene-text images. The MARIO-Eval benchmark contains 5,414 prompts. This benchmark is composed of six different subsets, each a subset of multiple benchmarks, including DrawBenchText \cite{drawtext}, DrawTextCreative \cite{drawtext}, ChineseDrawText \cite{glyphdraw}, and Mario-10M test set \cite{textdiffuser}. Following the protocol \cite{textdiffuser}, we exclude images generated from ChineseDrawText and DrawBenchText, hence 5,000 images are used for measuring FID.

\paragraph{Benchmark for multi-lingual generation.}
Due to the lack of pre-existing benchmarks for MLT image generation, we developed a new benchmark for model evaluation on nine different languages. We create 25 prompt templates, such as ``A raccoon holding a paper saying words.''. We exclude the non-English target words from the prompts, as they cannot be processed by the text encoder. Each prompt is designed to include up to three words in the image. Except for English, we allocate the rest of eight languages a set of 200 prompts, and 400 for English, yielding 2,500 prompts in total. For computing FID, we construct a set of images by randomly sampling a subset of the ICDAR2019 dataset.

\paragraph{Benchmark for logo generation.}
We additionally construct a benchmark for logo image generation. Specifically, we create 25 template prompts, e.g., ``A building displaying a [KEYWORD] sign.'', where [KEYWORD] here is the name of the logo. We obtain 150 logos in total, which is the union of the class of FlickrLogos-32 dataset \cite{flickr27} and LITW dataset, yielding 1,500 prompts in total. For computing FID, we construct an image set sampled from the merged set of FlickrLogos-32 \cite{flickr32} and LITW \cite{litw} dataset.


\subsection{Text Image Editing Details} 
We elaborate on the extension of our model for text image editing. In line with \cite{textdiffuser}, we augment the input channel by incorporating additional encoded latent obtained from masked images. Specifically, we use VAE to encode an image containing masked regions and concatenate the encoded latent with the original model input. We use the same VAE used for reference image encoding to encode the masked image, and noise is not added to the resulting latent. Consequently, we obtain an augmented input $\tilde{\modelinput}_{i}^t\in\mathcal{R}^{(3c+1)\times h \times w}$ as depicted below:
\begin{equation}
\small
    \tilde{\modelinput}_{i}^t=\text{concat}(\imglatent_{i}^t, \reflatent_{i}, \mask_{i},\imglatent_{i}^\text{masked}),
    \label{eq:input}
\end{equation}
where $\imglatent_{i}^{\text{masked}}\in\mathcal{R}^{c\times h\times w}$ denotes VAE encoded latent of the masked image. During training, along with the text region, we randomly select an additional 1-3 random regions to be masked. During inference, we mask the region to be modified. We provide additional qualitative results in \Cref{app:results}.

\section{Additional Ablation Studies}
\label{app:ablation}
In this section, we discuss additional ablation studies.

\subsection{Ablation Study on Languages}
\label{app:abl_lang}

\begin{table*}[h!]
\footnotesize
\begin{center}
  \resizebox{\linewidth}{!}{
  \begin{tabular}{cccccccccc|cccc}
    \toprule
    \textbf{Method}&\textbf{English}    &\textbf{Italian}& \textbf{German} &\textbf{French} &\textbf{Hindi} &\textbf{Bengali} &\textbf{Russian} &\textbf{Thai} & \textbf{Greek} &\textbf{Latin} & \textbf{Non-Latin} &\textbf{Mean} &\textbf{FID} \\
    \midrule
    ControNet&15.00&12.00&08.50&07.00&08.50&11.00&15.00&6.00&19.50&10.63&12.00&11.75&119.996    \\
    SD&0.00&0.50&0.0100&0.00&0.00&0.00&0.00&0.00&0.00&0.38&0.00&0.15&\textbf{93.198}    \\
    TextDiffuser&\textbf{84.00}&\textbf{59.00}&\textbf{62.50}&27.50&0.00&0.00&0.00&0.00&0.00&58.25&0.00&31.70&96.826    \\
    \midrule
    Ours&70.00&48.00&56.00&51.50&21.00&\textbf{11.50}&38.00&15.50&33.00&56.38&23.80&41.45 &97.523  \\
    Ours$^\dagger$&81.75&58.50&60.50&\textbf{57.50}&\textbf{22.00}&10.00&\textbf{48.00}&\textbf{18.00}&\textbf{48.00}&\textbf{64.56}&\textbf{29.20}&\textbf{48.60}&97.156    \\
    \bottomrule
  \end{tabular}
  }
  \caption{\textbf{OCR Accuracy comparison result for MLT image generation, presented individually for each language.} Results marked with $^\dagger$ indicate the utilization of CFG scheduling. We set $\rho_\text{speed}=0.2$, $\gamma=3.5$ and $\omega_\text{all}=4$ for CFG scheduling.}
    \label{tab:mlt_per_lang_comparison}
\end{center}
\vspace{-5mm}
\end{table*}

\begin{table}[h!]
\footnotesize
\begin{center}
  \resizebox{\linewidth}{!}{
  \begin{tabular}{ccccccc}
    \toprule
    \textbf{Script}&\textbf{Language}    &\textbf{FID}$\downarrow$   & \textbf{CLIP}$\uparrow$ &\textbf{Acc}$\uparrow$  &\textbf{F-1}$\uparrow$ \\
    \midrule
    \multirow{4}{*}{Latin}&English & 157.954 &0.2977 & 70.00    & 87.80    \\
    &Italian & 195.653 &0.2946 & 48.00    & 70.43    \\
    &German  & 195.230 &0.2918 & 56.00    & 78.47    \\
    &French  & 190.083 &0.2942 & 51.50    & 70.01    \\
    \midrule
    \multirow{5}{*}{Non-Latin}&Hindi   & 193.788 &0.3125 & 21.00    & 37.54    \\
    &Bengali & 192.716 &0.3135 & 11.50    & 17.60    \\
    &Russian & 188.567 &0.2998 & 38.00    & 56.89    \\
    &Thai    & 203.457 &0.3015 & 15.50    & 27.59    \\
    &Greek   & 192.215 &0.3023 & 33.00    & 55.20    \\
    \midrule
    &Average                  & 97.523 &0.3006 &41.45    &  58.94    \\ 
    \bottomrule
  \end{tabular}
  }
  \caption{\textbf{Evaluation results for MLT image generation, presented individually for each language.}}
  \label{tab:mlt_per_lang_eval}
\end{center}
\vspace{-5mm}
\end{table}

\paragraph{Per language comparison.}
We provide OCR Accuracy for each language and compare it with the existing methods in \Cref{tab:mlt_per_lang_comparison}. 
Remarkably, our model is shown to be the most accurate, as it surpasses existing methods across all average accuracy metrics, including the average of Latin, non-Latin, and the entire language set by a significant margin. Moreover, our method achieves high English accuracy comparable to that of TextDiffuser, which was tailored for English image generation. Although TextDiffuser shows high accuracy in some languages, such as English and German, it shows incapability in the rest of the language set, as it achieves 0 accuracy. In contrast to TextDiffuser, our method can generate text in all languages, and this demonstrates the multi-lingual capability of our model.
\smallbreak

\paragraph{Latin \textit{v.s.} non-Latin.} Our model exhibits higher OCR Accuracy scores in Latin languages such as English, Italian, German, and French (\Cref{tab:mlt_per_lang_comparison,tab:mlt_per_lang_eval}). It is noteworthy that for Russian, Thai, and Greek, our model is exclusively trained on synthetic images, yet it achieves comparable or superior FID and CLIP scores compared to the results for Latin languages. Given that the model was pretrained on a large-scale English dataset \cite{textdiffuser}, we speculate that this contributes as one factor to its superior performance in Latin languages.
\smallbreak

\paragraph{Different levels of complexity among languages.}
We note that different languages present different levels of complexity, with some being more challenging to generate accurately (\Cref{tab:mlt_per_lang_comparison,tab:mlt_per_lang_eval}). For instance, the size of the alphabet in English is 26, whereas in Thai, the size is 59. 
In addition, in languages such as Thai and Greek, diacritics are combined with other letters, and extra complexity is added.
Some of the predictions that our model makes look similar to the actual word, but are evaluated to be incorrect due to the mis-generation of diacritics. Besides, the shapes of characters in non-Latin languages (e.g., Bengali), are notably more intricate than Latin characters, whose alphabet typically requires a greater number of strokes to form.
\\

\begin{table*}
\footnotesize
\begin{minipage}{\linewidth}
    \begin{minipage}{0.5\linewidth}
            \centering
            \begin{tabular}{ccc|ccc}
            \toprule
            \textbf{$\rho_\text{speed}$} & $\gamma$ & $\omega_{\text{all}}$&  \textbf{FID}$\downarrow$ &\textbf{CLIP}$\uparrow$ &\textbf{ Acc}$\uparrow$ \\
            \midrule
            0.2 &  3.5  & 4 & 45.355 &0.3406 &\textbf{64.65}  \\
            0.5 &  3.5  & 4 & 44.680 &0.3419 &63.11  \\
            1.0 &  3.5  & 4 & \textbf{43.884} &\textbf{0.3428} &60.53  \\
            \bottomrule
            \end{tabular}
            \caption{\textbf{Evaluation results on MARIO-EVAL with varying CFG scheduling speeds $\rho_\text{speed}$.}} 
        \label{tab:abl_cfg_schedule_speed}
    \end{minipage}
    \hspace{0.01\linewidth}
    \begin{minipage}{0.5\linewidth}
            \centering
            \begin{tabular}{ccc|ccc}
            \toprule
            $\rho_\text{speed}$ & $\gamma$ & $\omega_{\text{all}}$&  \textbf{FID}$\downarrow$ &\textbf{CLIP}$\uparrow$ &\textbf{ Acc}$\uparrow$ \\
            \midrule
            0.5 &  0.5  & 7 & \textbf{38.104} &\textbf{0.3454} &59.84  \\
            0.5 &  1.5  & 6 & 39.932 &0.3448 &59.25  \\
            0.5 &  2.5  & 5 & 42.192 &0.3434 &61.73  \\
            0.5 &  3.5  & 4 & 44.680 &0.3419 &62.51  \\
            0.5 &  4.5  & 3 & 49.827 &0.3317 &\textbf{63.85}  \\
            \midrule
             - &  0  & 7.5 & 38.593 & \textbf{0.3454}& 58.26  \\
            \bottomrule
            \end{tabular}
        \caption{\textbf{Evaluation results on MARIO-EVAL with varying portions of scheduled guidance $\gamma$.} $\gamma=0$ denotes the result of the CFG without the guidance scheduling.}
        \label{tab:abl_cfg_schedule_scale}
    \end{minipage}
\end{minipage}
\end{table*}

\begin{figure*}[h!]
\begin{center}
    \includegraphics[width=0.85\textwidth]{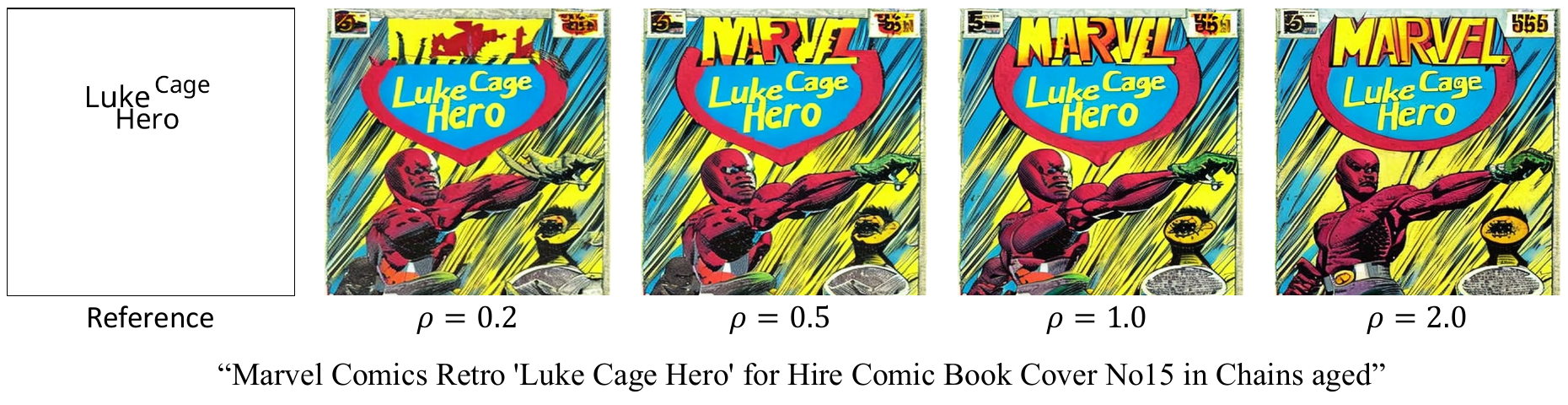}
    \caption{\textbf{Text-to-image generation with different CFG scheduling speeds.} The utilized reference image is denoted on the left where the target words to be generated are rendered. The text prompt with text words to be generated enclosed with punctuations is denoted at the bottom.
    Note the non-target word `MARVEL' starts to appear at the top of each image as $\rho_\text{speed}$ increases. We use identical random seeds for all the generations.}
    \label{fig:app_cfg_impact_speed}
    \vspace{-5mm}
\end{center}
\end{figure*}

\subsection{Ablation Study on CFG scheduling} 
We provide an ablation study on the proposed Augmented Classifier-Free Guidance with scheduling detailed in \Cref{app:detail_aug_cfg} in \Cref{app:detail_aug_cfg}. For better reference, we repeat the \Cref{eq:cfg_schedule_app} as below:
\begin{align*}
\small
\begin{split}
    \hat{\epsilon}^t & = \underbrace{\epsilon_\theta(\text{concat}(\imglatent^t, \varnothing, \varnothing),  \varnothing, t)}_\text{Unconditional} \\
    & \underbrace{+ \omega_{\text{prompt}} \Big[\epsilon_\theta(\text{concat}(\imglatent^t, \varnothing, \varnothing), \prompt, t) - \epsilon_\theta(\text{concat}(\imglatent^t, \varnothing, \varnothing), \varnothing, t)\Big]}_\text{Caption Guidance} \\
    &+ \underbrace{\omega_{\text{ref}}\Big[\epsilon_\theta(\text{concat}(\imglatent^t,\textbf{s},\textbf{m}), \varnothing, t) - \epsilon_\theta(\text{concat}(\imglatent^t, \varnothing, \varnothing), \varnothing, t)\Big]}_\text{Reference Guidance} \\
    &+ \underbrace{\omega_{\text{all}}\Big[\epsilon(\text{concat}(\imglatent^t,\textbf{s},\textbf{m}), \prompt, t) - \epsilon(\text{concat}(\imglatent^t, \varnothing, \varnothing), \varnothing, t)\Big]}_\text{All Guidance},
\end{split}
\end{align*}
where $\omega^t_\text{ref}$ and $\omega^t_\text{prompt}$ are the values of guidance scale for reference and prompt condition at timestep $t$, and $\omega_\text{all}$ is a guidance scale for both conditions which is set to be a constant. We dynamically schedule the guidance scales as follows:
\begin{align*}
    \omega^t_\text{ref} &= \gamma{\frac{t}{\bar{T}}}^{\rho_\text{speed}} \\
    \omega^t_\text{prompt} &= \gamma(1-{\frac{t}{\bar{T}}}^{\rho_\text{speed}}),
\end{align*}
where $t$ is counted in decreasing order, hence $\omega_{\text{ref}}$ is scheduled to be decreasing, and $\omega_{\text{prompt}}$ to be increasing. Here, $\gamma$ is a constant value corresponding to the max/min value of $\{\omega^t_\text{r}\}$ and $\{\omega^t_\text{c}\}$, and $\rho_\text{speed}$ denotes the speed of each guidance schedule.

\smallbreak
\paragraph{Impact of the scheduling speed.}
We first study the impact of CFG scheduling by varying $\rho_\text{speed}$, the increase/decrease rate of the guidance scale, in \Cref{tab:abl_cfg_schedule_speed}. We observe a rapid decrease in $\omega_\text{ref}$ with a rapid increase in $\omega_\text{ref}$ result in a decrease in the OCR Accuracy and improvements in FID and CLIP scores. As $\omega_\text{ref}$ decreases rapidly, the impact of the prompt gets higher and the reference condition becomes less influential in the generation process. As a consequence, non-target words within the prompt gain increased influence and affect the generation, resulting in the inclusion of non-target words in the output (\Cref{fig:app_cfg_impact_speed}).
\smallbreak
\paragraph{Impact of different portions of scheduled guidance.}
We examine the effect of different portions of scheduled guidance $\gamma$ within the total amount of guidance (\Cref{tab:abl_cfg_schedule_scale}). The result of CFG without scheduling is also provided for comparison. We observe the general trend of a decrease in CLIP scores and an increase in OCR Accuracy as $\gamma$ increases. 
We speculate that the initial guidance scale of the reference condition plays an important role in establishing the initial layout of the text to be generated which, in turn, influences the effect of the prompt condition and consequently affects CLIP and FID.

\subsection{Ablation Study on Hyper Parameters}
\paragraph{Impact of synthetic loss weight.} We first analyze the impact of the synthetic loss weight in \Cref{tab:abl_synth_loss_weight}. We note that the synthetic loss weight mainly impacts the \textbf{CLIP} as the text prompts of the real dataset better describe the content of an image. We observe the model trained with a synthetic loss weight of 0.5 achieves a lower \textbf{CLIP} than the one trained with $1.0$, which validates the effectiveness of lowering the weight of reconstruction loss when synthetic images are provided.
\smallbreak

\paragraph{Impact of recognition loss weight.} We analyze the impact of the recognition loss weight on accuracy and the quality of the generated results in \Cref{tab:abl_rec_loss_weight}. 
We note that the OCR accuracy improves as the model is trained with increased recognition loss weight. In general, models with higher loss weight achieve better OCR Accuracy and a worse score on FID. We choose the recognition weight as $0.025$ for a proper trade-off.
\smallbreak

\begin{table}[h!]
\footnotesize
\begin{minipage}{\linewidth}
\begin{center}
    \begin{minipage}{0.42\linewidth}
    \centering
    \begin{tabular}{c|c}
        \toprule
        \textbf{Weight} & \textbf{CLIP}$\uparrow$ \\
        \midrule
        0.5   &0.3011   \\
        1.0    &0.2997 \\
        \bottomrule
    \end{tabular}
    \caption{\textbf{Ablation study on synthetic loss weight for multi-lingual image generation on Russian subset.}} 
    \label{tab:abl_synth_loss_weight}
    
    \end{minipage}
    \hspace{0.005\linewidth}
    \begin{minipage}{0.45\linewidth}
    \footnotesize
        \centering
        \begin{tabular}{c|cc}
            \toprule
            \textbf{Weight} & \textbf{Acc}$\uparrow$ & \textbf{FID}$\uparrow$ \\
            \midrule
            0.025   &43.47 & 89.97  \\
            0.05    &46.13 & 92.82 \\
            \bottomrule
        \end{tabular}
        \caption{\textbf{Ablation study on recognition loss weight for logo image generation.}  Results are denoted in [\%].}
        \label{tab:abl_rec_loss_weight}
    \end{minipage}
\end{center}
\end{minipage}
\end{table}

\begin{table}[t!]
\footnotesize
\centering
\begin{tabular}{cccc}
\toprule
\textbf{Rank} & \textbf{F-1} $\uparrow$ & \textbf{CLIP}$\uparrow$ &\textbf{\# Lora Params (M)}$\downarrow$\\
\midrule
1      & 51.58    & 0.3359   & 0.89   \\
4      & 69.44    & 0.3372   & 3.57   \\
\textbf{32}    &\textbf{70.52 }   & \textbf{0.3445}   & \textbf{28.55} \\
128  &70.19   & 0.3436   & 114.21 \\
\bottomrule
\end{tabular}
\caption{\textbf{Ablation study for different rank settings of LoRA \cite{lora}.} The model is trained on MARIO-LAION \cite{textdiffuser}. The number of parameters indicated on a million scale. We denote with \textbf{bold} for the selected rank configuration.}
\label{tab:abl_lora}
\end{table}

\paragraph{Impact of LoRA rank.} We examine the influence of various rank configurations for LoRA \cite{lora} and provide justification for the selected rank. We assess models trained with different rank values trained with the MARIO-10M dataset (\Cref{tab:abl_lora}). We observe a consistent pattern of performance improvement with increasing rank values, then saturates at $r=32$, thus we opt for a rank setting of $r=32$ as our final choice.

\begin{table}[t!]
\footnotesize
\centering
\begin{tabular}{cccc}
\toprule
\textbf{Type} & \textbf{Acc} $\uparrow$ & \textbf{F1} $\uparrow$ & \textbf{CLIP}\\
\midrule
Blank Reference          & 20.27     & 27.18  & 0.3695 \\
Ours                             & \textbf{42.87}    & \textbf{48.86}  &\textbf{0.3759} \\
\bottomrule
\end{tabular}
\caption{\textbf{Impact of subject existence in logo generation.}}
\label{tab:abl_blank_reference_logo}
\end{table}

\begin{table}[t!]
\footnotesize
\centering
\begin{tabular}{cccc}
\toprule
\textbf{Type} & \textbf{Acc} $\uparrow$ & \textbf{F1} $\uparrow$ & \textbf{CLIP}\\
\midrule
Blank Reference          & 1.88     & 2.71  & 0.2855 \\
Ours                             & \textbf{55.19}    & \textbf{79,24}  &\textbf{0.3468} \\
\bottomrule
\end{tabular}
\caption{\textbf{Impact of subject existence in English generation.}}
\label{tab:abl_blank_reference_english}
\end{table}

\begin{table}[h]
\footnotesize
\centering
    \begin{tabular}{c|cc}
        \toprule
        Model & \textbf{Acc}$\uparrow$ &\textbf{CLIP}$\uparrow$ \\
        \midrule
        SD-v1.5 & 0.03 & 0.3015\\
        Ours-v1.5 & \textbf{51.40} & \textbf{0.3647}\\
        \midrule
        SD-v2.1 & 0.02          & 0.3221 \\
        Ours-v2.1 & \textbf{49.20}         & \textbf{0.3685}\\
        \bottomrule
    \end{tabular}
    \caption{\textbf{Evaluation of different SD versions on the OpenLibraryEval500 subset.}}
    \label{tab:app_sd_versions}
\end{table}

\begin{table}[h]
\centering
\footnotesize
    \begin{tabular}{c|c}
    \toprule
    Training Task & \textbf{Acc}$\uparrow$  \\
    \midrule
    English        &\textbf{58.26}\\
    English+Logo   & 51.64   \\
    \bottomrule
    \end{tabular}
    \caption{\textbf{Simultaneous training result for English generation.}}
    \label{tab:app_sim_english}
    \vspace{-2mm}
\end{table}


\subsection{Analysis on Reference Image.}
\begin{figure*}[t]
    \centering
    \includegraphics[width=\textwidth]{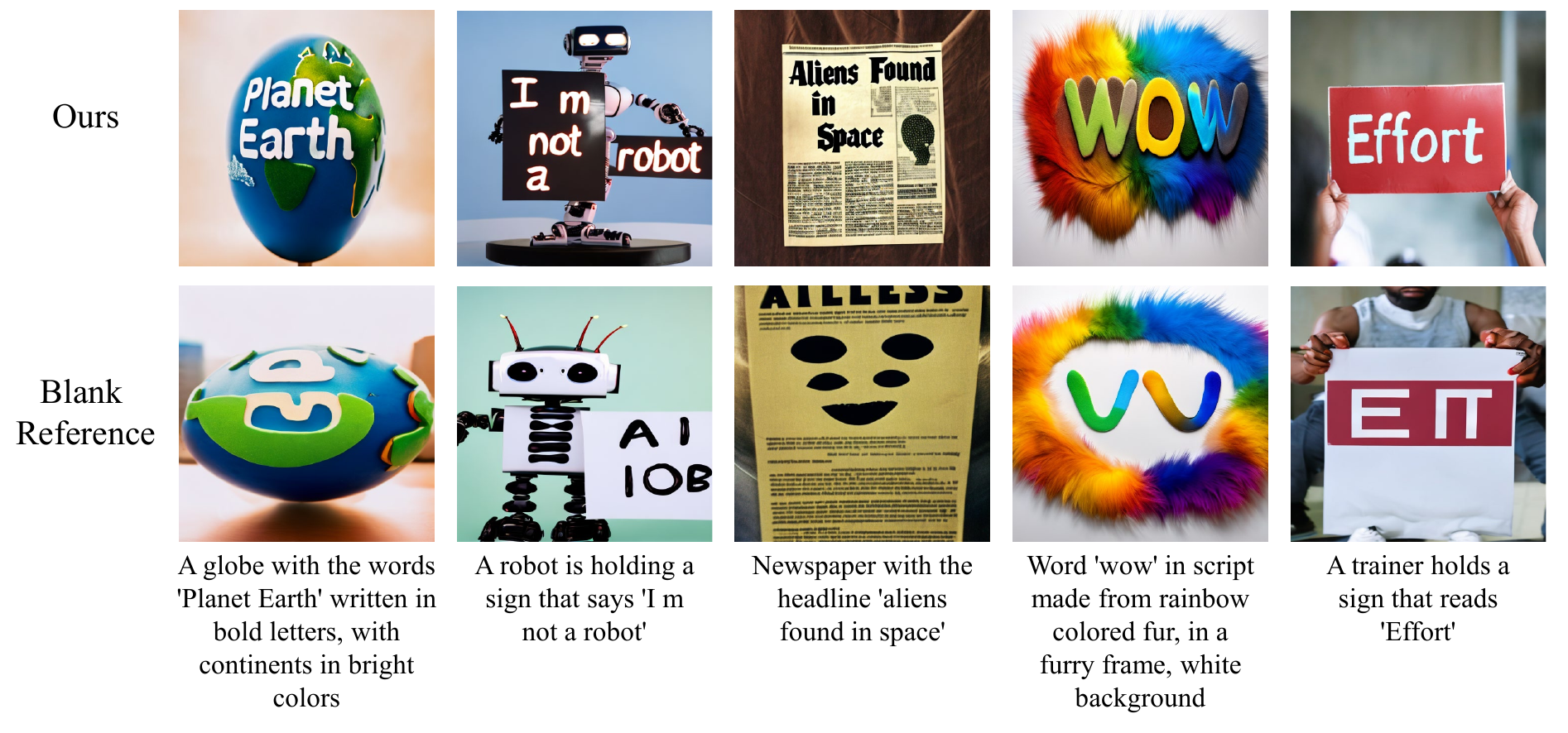}
    \vspace{-5mm}
    \caption{\textbf{Generation result with blank reference image.}}
    \label{fig:app_ref_abl}
\end{figure*}

\paragraph{Impact of reference image during inference.}
To study the impact of the appearance of the reference image, we compare generation results using the reference image with and without subject rendering \textit{i.e.,} blank reference in \Cref{tab:abl_blank_reference_logo,tab:abl_blank_reference_english}. The results show that the generation result is greatly influenced by the appearance of the image. We also provide qualitative results in \Cref{fig:app_ref_abl}.

\subsection{Ablation Study on SD Versions}
To show generalization over different versions, we provide results on SD-v1.5 in \Cref{tab:app_sd_versions}. Our method effectively improves on both versions.

\subsection{Ablation Study on Simultaneous Task Training}
Our method employs separate plugin modules, each acting as an expert for a specific task. When a task is prompted, its corresponding expert plugin is loaded. This modular design reduces interference between tasks, which often occurs when training with diverse objectives. Experimental results confirm this advantage: simultaneous task training leads to degraded performance (\Cref{tab:app_sim_english}).

\begin{figure*}[t!]
    \centering
    \includegraphics[width=\textwidth]{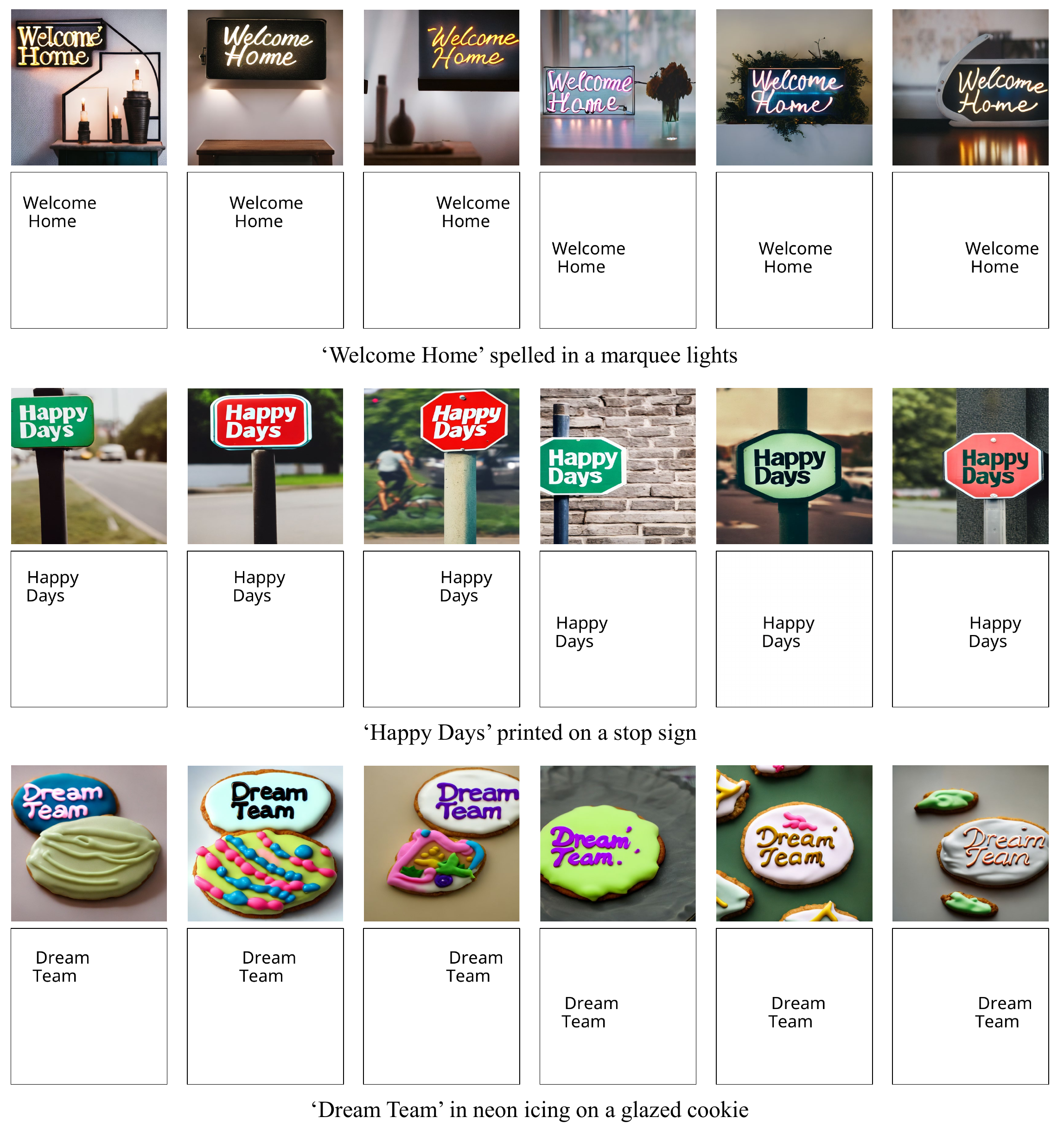}
    \caption{\textbf{Generation result with varying position.}}
    \label{fig:app_position_control}
\end{figure*}

\paragraph{Impact of subject location in reference.}
We present a visual analysis of how the spatial position of the subject in the reference image affects the generated outputs. As illustrated in \Cref{fig:app_position_control}, the subject’s location in the reference image is consistently reflected in its position within the generated results.

\begin{figure*}[t!]
    \centering
    \includegraphics[width=\textwidth]{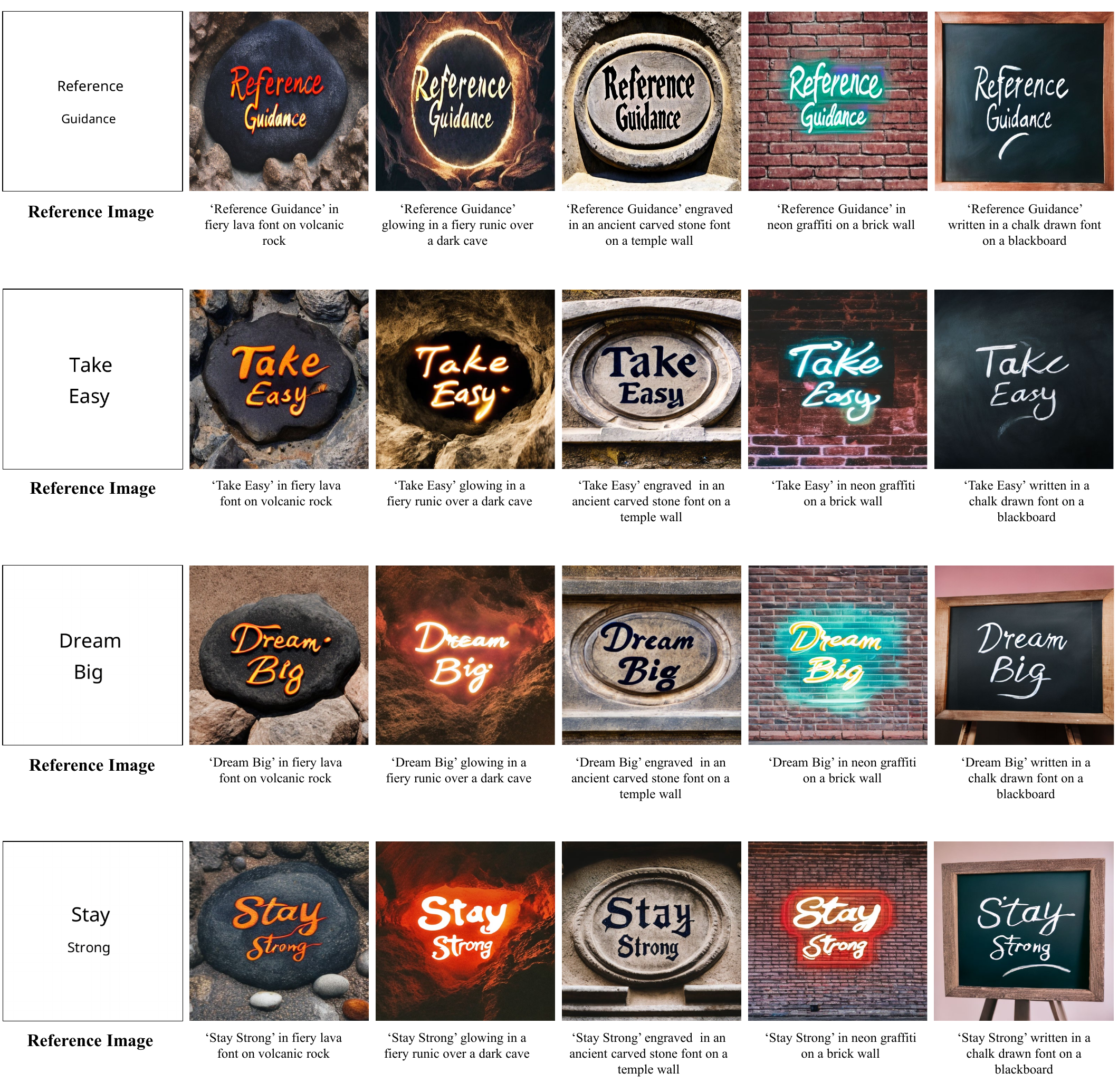}
    \caption{\textbf{Generation result with the same reference image.} The images in the same row are generated with the same reference, but with different input prompts.}
    \label{fig:app_style_control}
\end{figure*}

\paragraph{Joint influence of prompt and reference.} We observe that \emph{both} the input prompt and the reference image influence the generation results. To show this, we visualize the generation result with the same reference image, but with a different input prompt. As shown in \Cref{fig:app_style_control}, the generation results not only depend on the appearance of the subject in the reference image, but also on the input prompt.

\section{Additional Qualitative Results}
\label{app:results}
In this section, we provide additional qualitative results.
\vspace{2mm}

\begin{figure*}[h!]
    \centering
    \includegraphics[height=0.98\textheight]{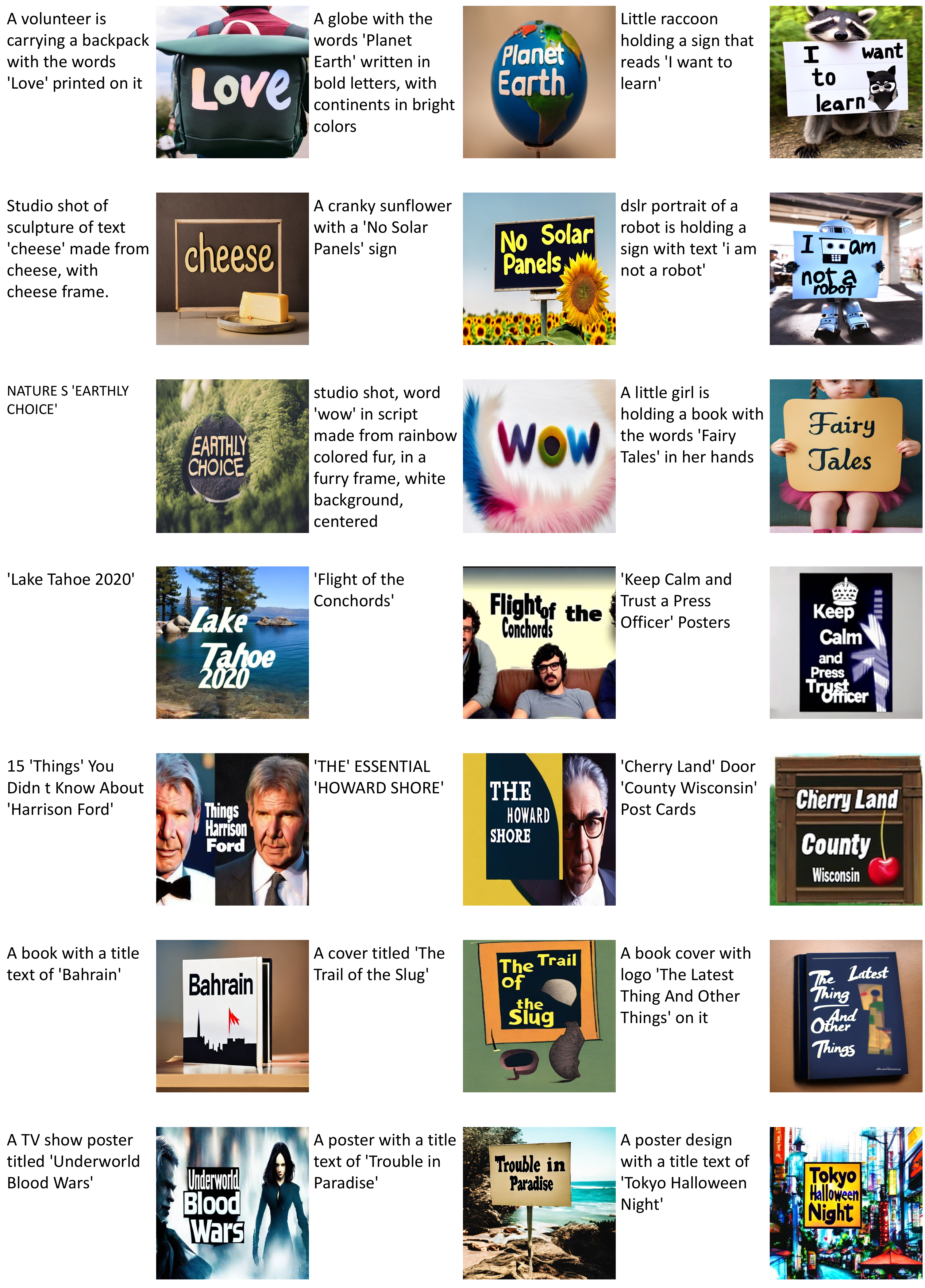}
    \caption{\textbf{Additional results for English generation.} The odd columns are the prompts, and the even columns are the generated images. Words enclosed with punctuation are the target keywords to be generated.}
    \label{fig:app_english}
\end{figure*}

\paragraph{Additional qualitative results for English texts.} 
We provide additional qualitative results for English text image generation (\Cref{fig:app_english}). The model fine-tuned on AText is adopted for the English image generation. We use the MARIO-EVAL benchmark prompts \cite{textdiffuser} for the generation.

\begin{figure*}[h!]
    \includegraphics[width=\textwidth]{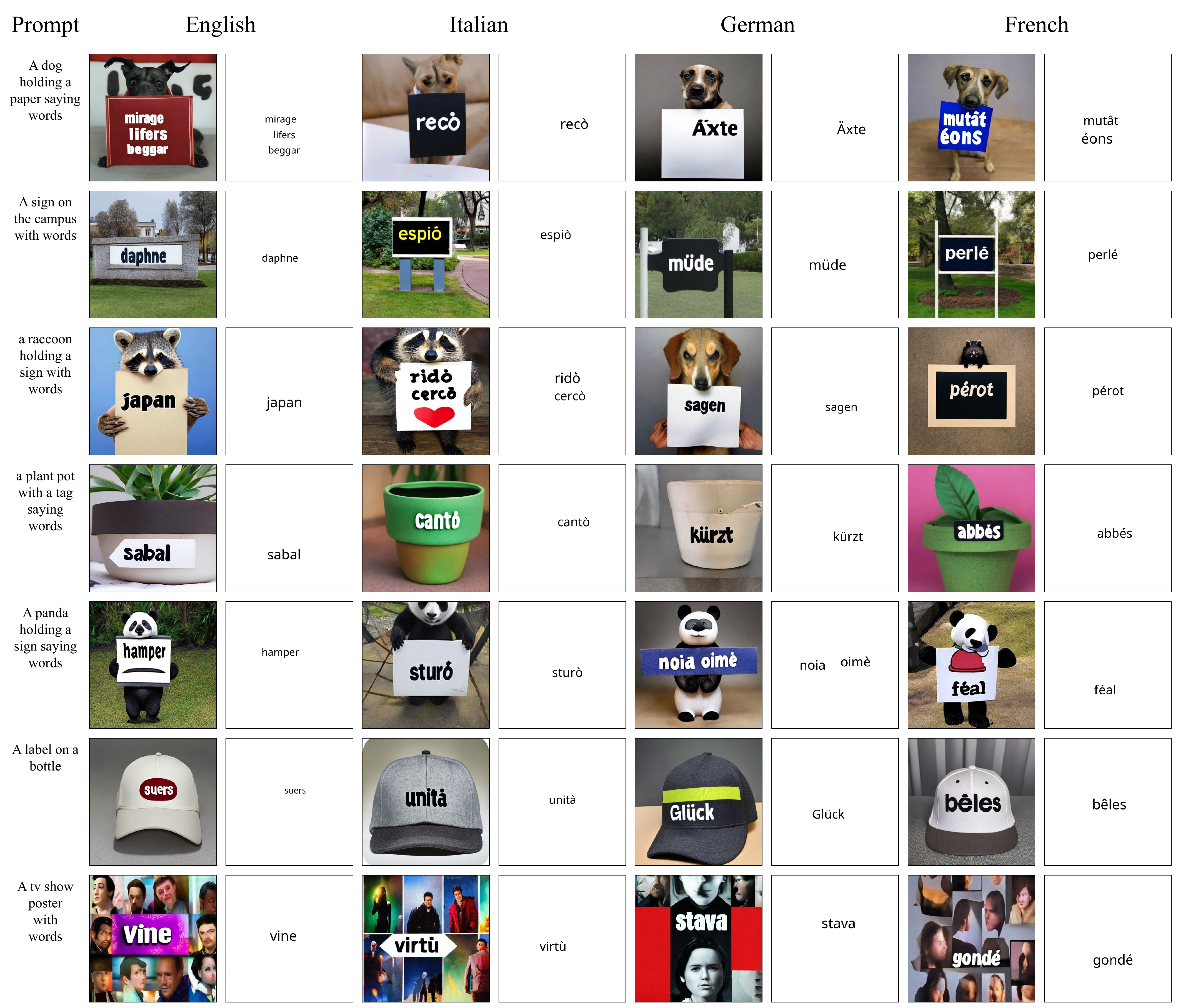}
    \caption{\textbf{Additional qualitative results for multi-lingual, Latin text image generation.} Images in the same row correspond to the same prompt templates. The first column denotes the prompts. The even column denotes the generated results, and the odd column denotes the reference image of text renderings.}
    \label{fig:app_mlt_latin}
\end{figure*}

\begin{figure*}[h!]
    \centering
    \includegraphics[width=\textwidth]{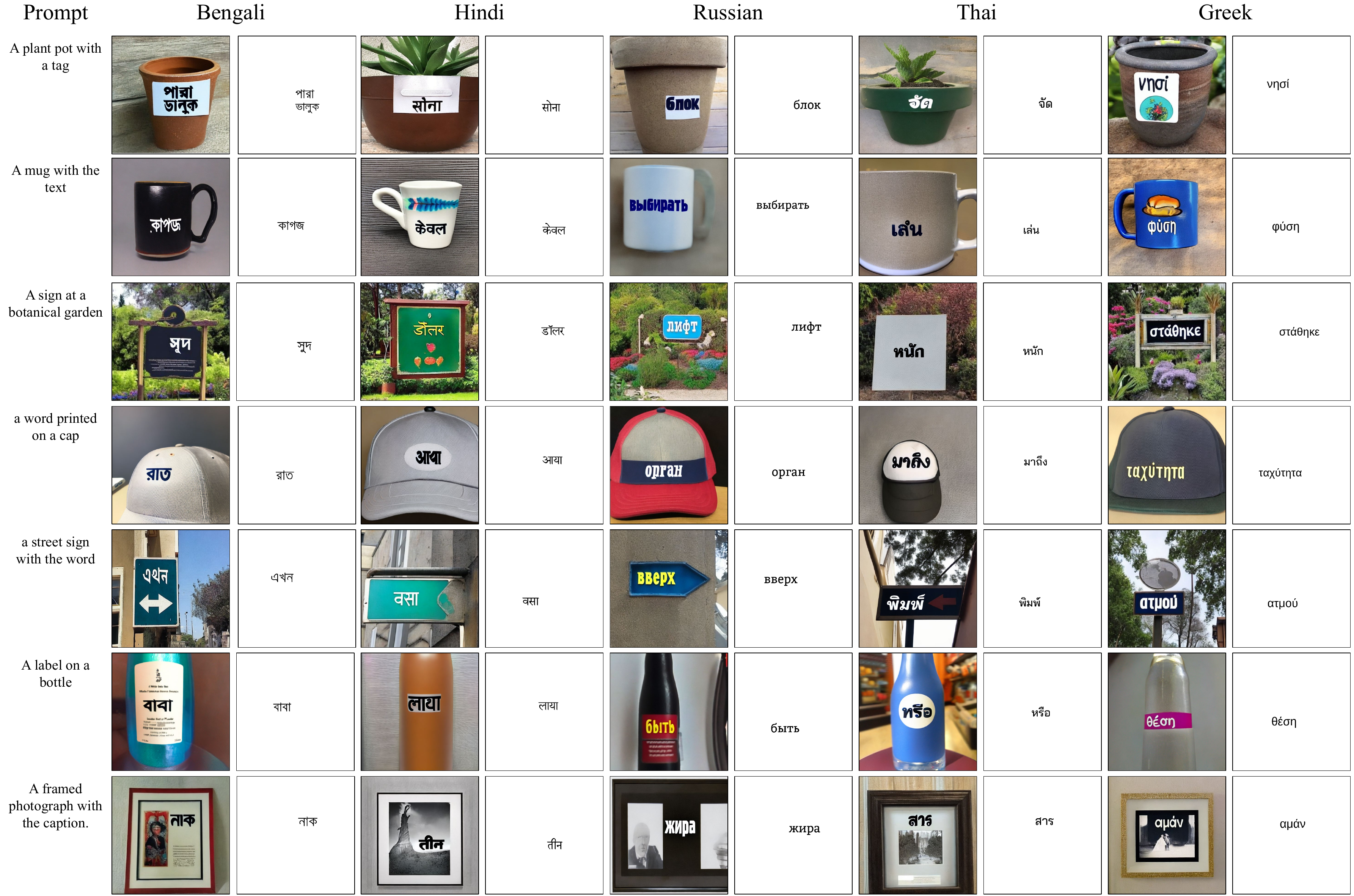}
    \caption{\textbf{Additional qualitative results for multi-lingual, non-Latin text image generation.} Images in the same row correspond to the same prompt templates. The first column denotes the prompt template. The even column denotes the generated results, and the odd column denotes the reference image of text renderings.}
    \label{fig:app_mlt_non_latin}
\end{figure*}

\paragraph{Additional qualitative results for multi-lingual texts.}
We provide additional qualitative results for the multi-lingual, Latin text image generation (\Cref{fig:app_mlt_latin}), and multi-lingual, non-Latin text image generation (\Cref{fig:app_mlt_non_latin}). We use the model trained on the merged set of ICDAR2019 \cite{icdar2019}, and synthetic images of all the languages. Despite the deficiency of real samples of Russian, Greek, and Thai, our model shows the capability of generalizing to these languages.

\begin{figure*}[h!]
\centering
    \includegraphics[width=\textwidth]{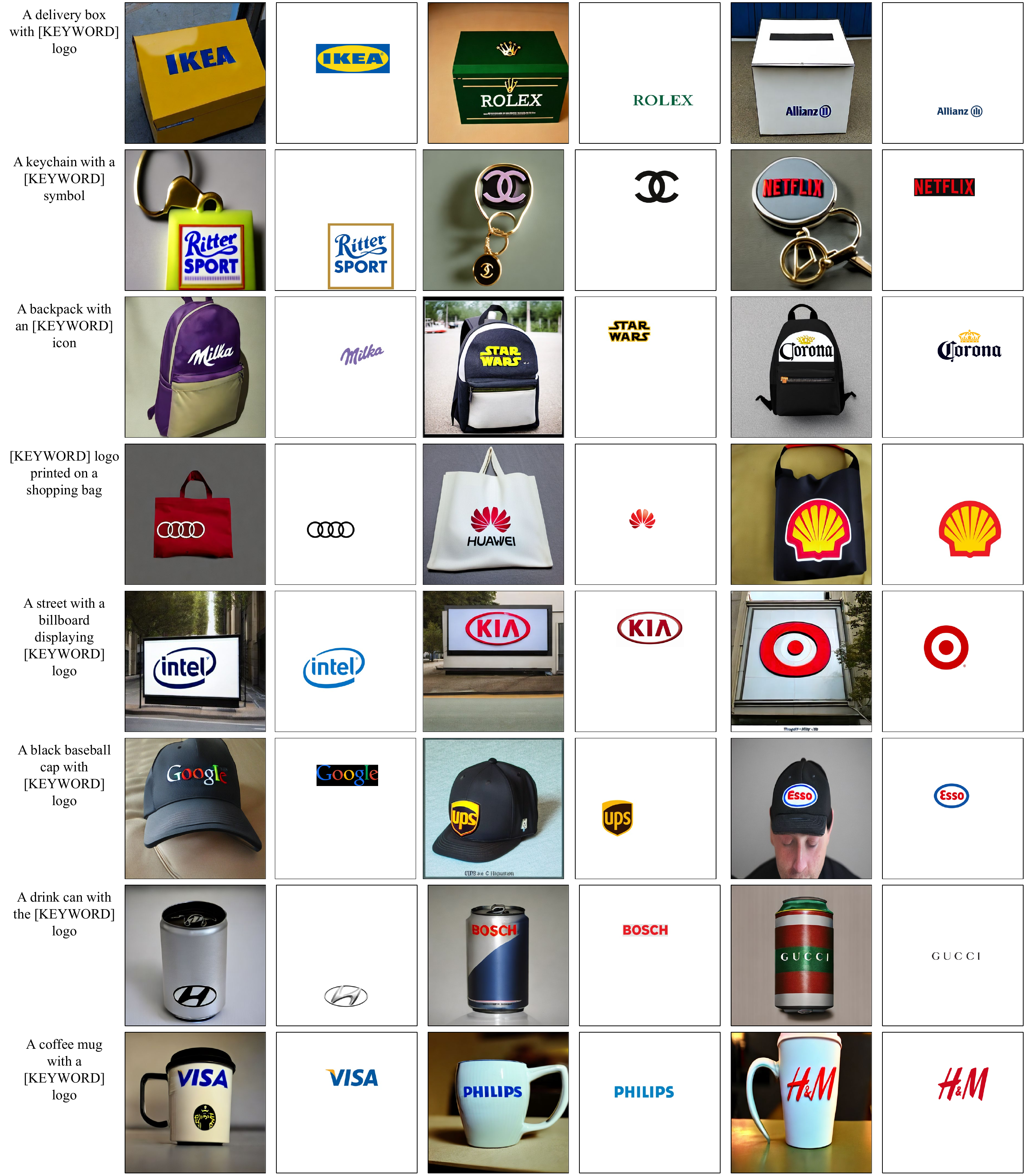}
    \caption{\textbf{Additional qualitative results for logo image generation.} The reference logos utilized here are part of the training logo set. Images in the same row correspond to the same prompt templates. The first column denotes the prompt template. We replace ``[KEYWORD]'' with the name of the logo during the inference. The even columns denote the generated results, and the odd columns denote the reference logo images.}
    \label{fig:app_seen_logos}
\end{figure*}

\begin{figure*}[h!]
    \centering
    \includegraphics[height=0.965\textheight]{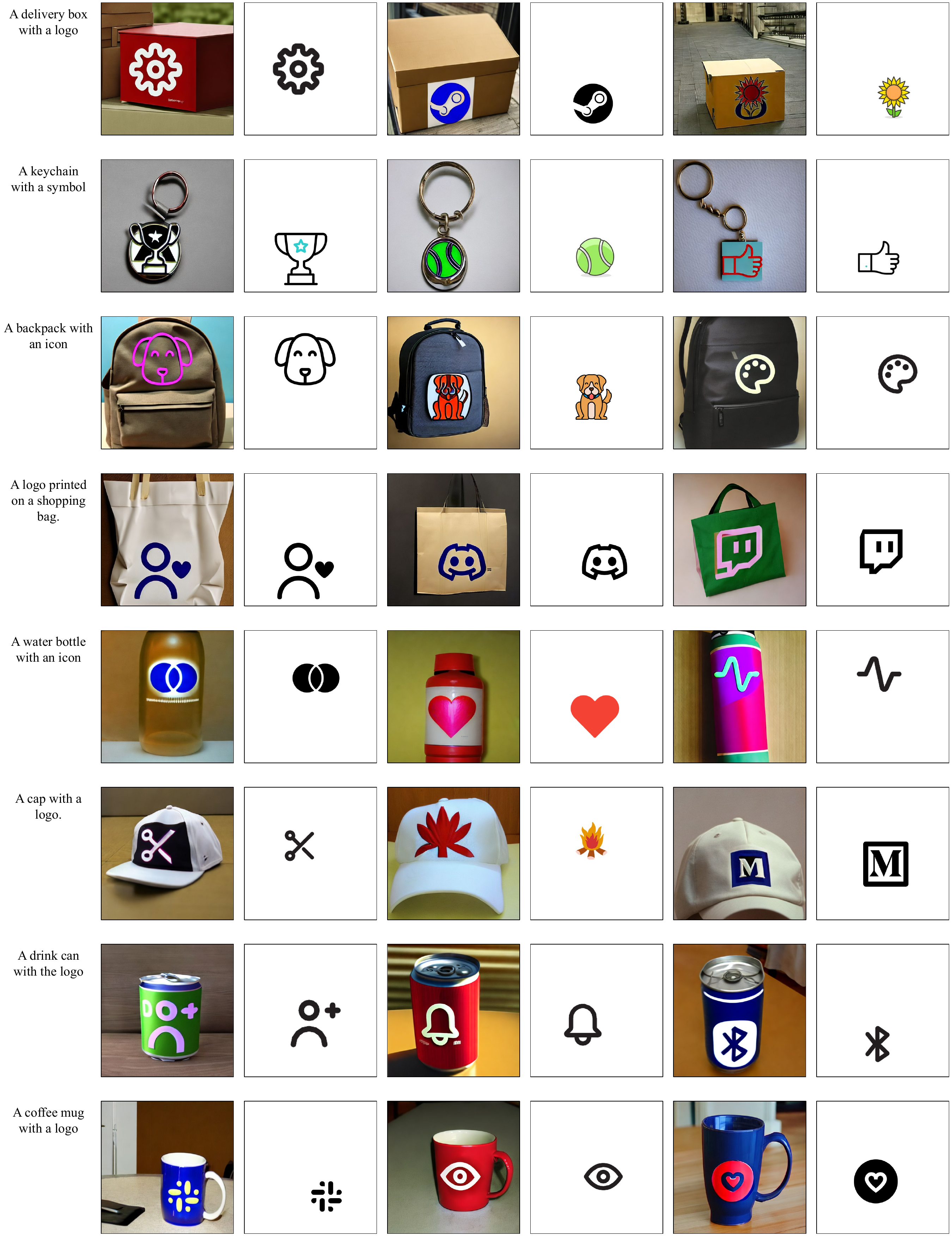}
    \caption{\textbf{Additional qualitative results for logo image generation.} The reference logos utilized here are not included in the training logo set and hence, are unseen during the model training. Images in the same row correspond to the same prompt templates. The first column denotes the prompt template. The even columns denote the generated results, and the odd columns denote the reference logo images.}
    \label{fig:app_unseen_logos}
\end{figure*}

\paragraph{Additional qualitative results for logos.} 
We present additional qualitative results of logo image generation. We first present the generation results of the logos that are included in the training set (\Cref{fig:app_seen_logos}). The results are generated using the logo benchmark prompts. The generated results confirm the model's capability of generating logos in the desired location of the corresponding reference logo image, and that this reference extends beyond text renderings. Moreover, we present the generation results of the logos that are never seen during training (\Cref{fig:app_unseen_logos}). The results further validate the effectiveness of the proposed method and the model's ability to generalize across novel instances.

\begin{figure*}[h!]
    \centering
    \includegraphics[width=\textwidth]{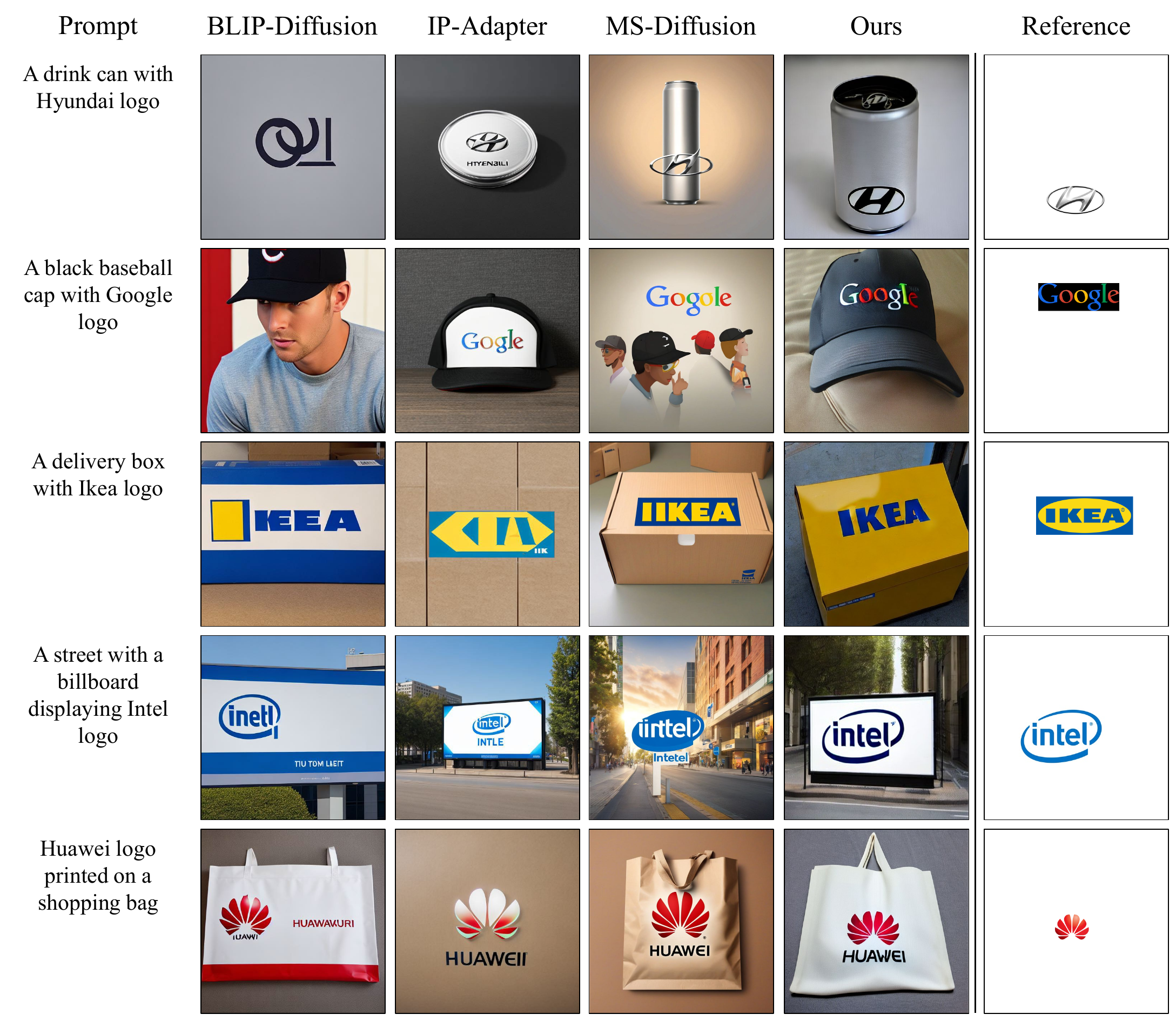}
    \caption{\textbf{Comparison results with personalization methods.}}
    \label{fig:app_personal_logo}
\end{figure*}

\paragraph{Comparison results with personalization methods.}
We provide comparison results with personalization methods in \Cref{fig:app_personal_logo}.

\begin{figure*}[h!]
    \centering
    \includegraphics[height=0.9\textheight]{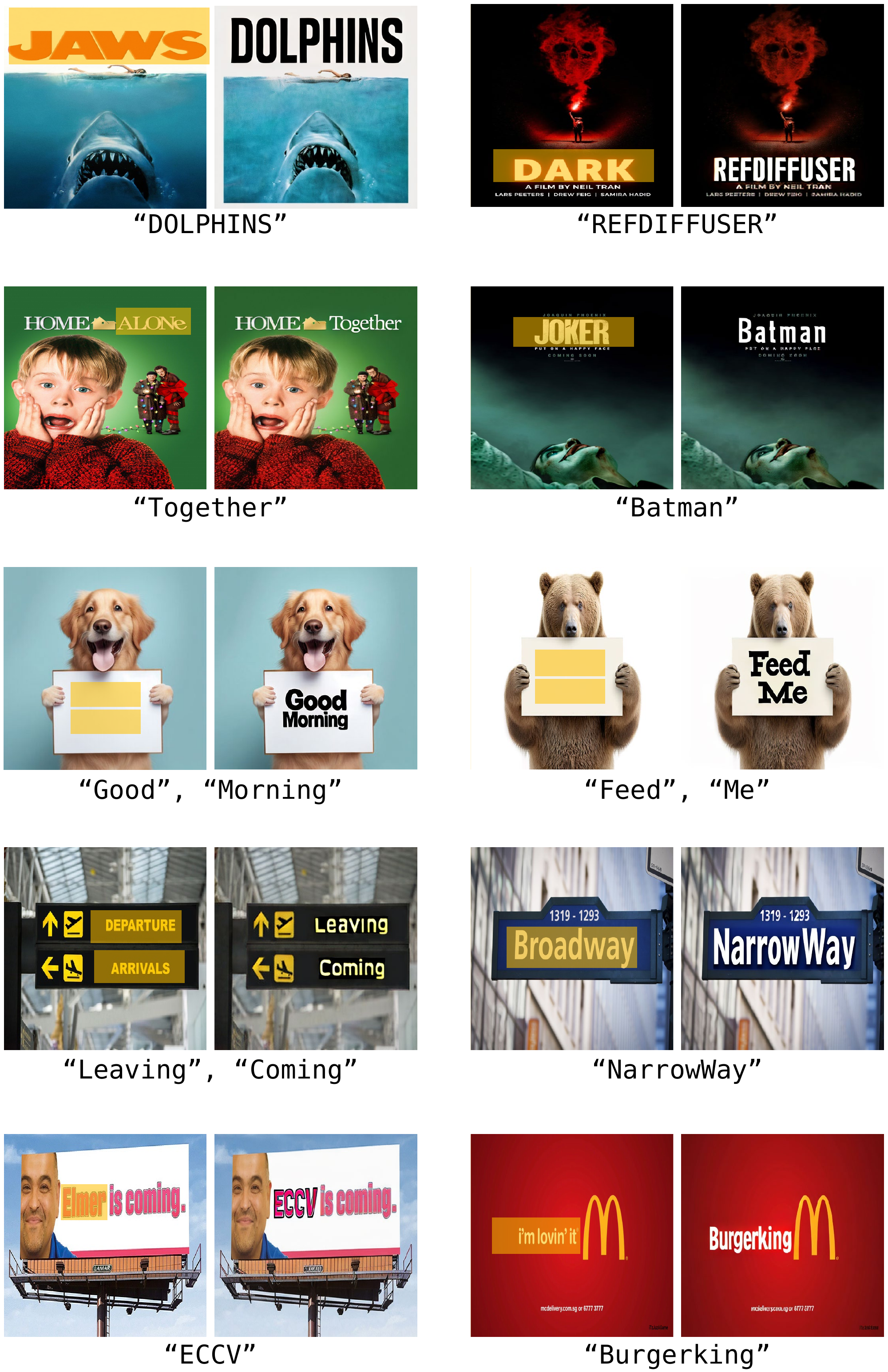}
    \caption{\textbf{Additional qualitative results for Text Image Editing.} Regions denoted in yellow are masked during the inference. We denote the added text below each of the results.}
    \label{fig:app_edit}
\end{figure*}

\paragraph{Additional text image editing results.}
We provide additional text image Editing results in \Cref{fig:app_edit}. The model successfully edits the specified region to include the intended words without modifying the rest of the region.


\clearpage


\end{document}